\documentclass[letterpaper]{article}
\usepackage[preprint]{aaai2027} 
\usepackage[hyphens]{url}         
\usepackage{graphicx}             
\usepackage{natbib}               
\usepackage{caption}              
\usepackage{amsmath}
\usepackage{amssymb}
\usepackage{amsthm}
\usepackage{mathtools}

\usepackage{booktabs}
\usepackage{tikz}
\usepackage{algorithm}
\usepackage{algorithmic}

\newtheorem{theorem}{Theorem}
\newtheorem{lemma}{Lemma}
\newtheorem{definition}{Definition}

\title{Zonotopic Analysis of Quantization Error in Neural Networks}

\author{
    Robert Debbas\textsuperscript{\rm 1},
    Eric Goubault\textsuperscript{\rm 2},
    Sylvie Putot\textsuperscript{\rm 2}
}

\affiliations{
    \textsuperscript{\rm 1}Harvard University\\
    \textsuperscript{\rm 2}LIX, École Polytechnique,
    Institut Polytechnique de Paris\\
    robertdebbas@g.harvard.edu,
    goubault@lix.polytechnique.fr,
    putot@lix.polytechnique.fr
}

\begin{document}
\maketitle

\begin{abstract}
    Quantized neural networks achieve computational efficiency by reducing precision, but bounding their quantization error remains a challenge due to discrete rounding and clamping operations. We present a zonotope-based abstract interpretation framework for quantized ReLU feed-forward network verification, introducing a unified round-clamp abstract transformer that jointly handles both operations. We extend this framework to constrained zonotopes, which add linear constraints to tighten bounds, and derive sound quantization error representations. We establish soundness and affine-strip optimality
    of the round-clamp abstract transformer, as well as convex-hull optimality of its complete constrained formulation. On the ACAS Xu benchmarks, our constrained zonotope method reduces average error-bound widths by approximately 70-85\% and runtimes by 65-80\% compared to QEBVerif's symbolic DRA propagation at the two smallest tested perturbation radii ($r=3$ and $r=6$). These improvements are not uniform across bitwidths, and symbolic DRA achieves tighter bounds at larger radii. Our plain zonotope method delivers competitive sound bounds nearly 300 times faster on average.
\end{abstract}

\section{Introduction}

Deep neural networks excel at complex tasks but require reduced complexity for deployment in constrained environments. Quantization addresses this by lowering parameter precision, yet introduces numerical errors that compromise its reliability in safety-critical applications like autonomous driving and medical diagnosis~\cite{song2023qnnrepairquantizedneuralnetwork}.

Verifying the quantized neural network fidelity requires computing formal bounds on quantization error: the deviation between original and quantized network outputs over input regions. This differs fundamentally from robustness verification~\cite{huang2017safety,katz2017reluplex}, which examines network behavior under input perturbations rather than quantization effects. While neural network robustness verification is mature~\cite{liu2021survey}, quantized network verification remains limited. Recent exact methods using integer linear programming~\cite{Zhang2022,huang2024efficient} and SMT solving~\cite{henzinger2021scalable,amir2021smt} are sound and complete but computationally expensive and struggle to scale.

The current state-of-the-art, QEBVerif~\cite{Zhang2023}, implements differential reachability analysis (DRA) with interval abstractions as its fast first stage, followed by MILP refinement. QEBVerif also provides a symbolic DRA variant based on polyhedral abstractions, which offers a natural point of comparison with our approach due to its ability to capture linear dependencies beyond intervals. In addition, both methods provide sound single-pass verification without refinement, which easily allows us to evaluate whether zonotopes improve precision and speed over symbolic DRA for quantization error analysis.

Zonotopes are center-symmetric convex polyhedra whose implicit encoding of linear dependencies enables efficient propagation through affine transformations while explicitly tracking correlations with noise symbols~\cite{Gehr2018,singh2018fast}. Zonotopes were shown to be very effective at propagating uncertainties and rounding errors in abstraction-based program analysis~\cite{vmcai11}. Differential zonotopes have also been used to track relationships between two neural networks~\cite{Teuber2025}, whereas our approach propagates the original and quantized networks separately. However, their adaptation to the evaluation of the quantization error in quantized networks, particularly for discrete rounding and clamping operations, remains unexplored. Constrained zonotopes~\cite{scott2016constrained} further tighten bounds by adding linear equality constraints. In this work, we use inequality constraints to partially lift the symmetry imposed by zonotopes and refine the abstraction of activations. Nonetheless, adding these constraints requires solving linear programming problems and scales up the cost of analysis, we thus investigate both the unconstrained and constrained cases.

Our results show that the plain zonotopic abstraction is on average 300 times faster than the symbolic DRA while still offering reasonable bounds, and the constrained zonotopic abstraction reduces average error-bound widths by approximately 70-85\% and runtimes by 65-80\% compared to symbolic DRA at the two smallest tested radii ($r=3$ and $r=6$). These improvements are not uniform across bitwidths, and DRA-Sym achieves tighter bounds at larger radii.

\textbf{Contributions.} To our knowledge, we present the first zonotope-based framework for quantization error bound verification of neural networks: (1) a novel abstract transformer: a unified round-clamp abstraction that captures both rounding and clamping with a single noise symbol, applicable to both standard and constrained zonotopes, (2) a complete zonotopic framework to compute formal bounds on the quantization error, and (3) soundness proofs and affine-strip width optimality for the round-clamp abstract transformer, together with convex-hull optimality of the complete constrained formulation, proved in
Appendices~\ref{app:soundness_optimality}
and~\ref{app:constrained_soundness}.

\section{Background}

To establish the theoretical foundation for our approach, we first review feed-forward neural network quantization and abstract interpretation. We present the quantization formalism necessary to understand how precision reduction affects network computations, followed by the abstract interpretation framework that enables efficient verification. We adopt notation from \cite{nagel2021whitepaperneuralnetwork} for quantization, \cite{Zhang2023} for quantized feed-forward neural networks (QNNs) and \cite{Gehr2018,Lemesle2024} for abstract interpretation.

\subsection{Neural Network Quantization}
\textbf{Quantization.} Neural network quantization maps real-valued network parameters and activations to a finite set of discrete values, typically represented as low-bit integers. We denote by $\mathbf{V}$ a tensor with real-valued coefficients (represented in practice as floating-point numbers) and by $\widehat{\mathbf{V}}$ its quantized counterpart. The quantized tensor is represented by a tensor $\mathbf{V}_{\text{int}}$ with integer coefficients and a scale factor $s$ that maps integers back to the real domain.

\textbf{Uniform quantization.} The quantization process maps each real value to its nearest representable integer value within a bounded range. Formally, uniform quantization is defined as:
$\mathbf{V}_{\text{int}} = \text{clamp}(\lfloor \mathbf{V}/s \rceil + z;\, C^{\mathrm{lb}}, C^{\mathrm{ub}})$,
where $s > 0$ is the scale factor, $z$ is the zero-point offset, and $C^{\mathrm{lb}}, C^{\mathrm{ub}}$ define the representable integer range. For symmetric uniform quantization, we set $z = 0$. The quantized tensor is then recovered as
$\widehat{\mathbf{V}} = s \cdot (\mathbf{V}_{\text{int}} - z)$.

\textbf{Quantized Neural Networks.} A feed-forward deep neural network $\mathcal{N} : \mathbb{R}^d \to \mathbb{R}^s$ with $N$ layers is $\mathcal{N} = l_N \circ \cdots \circ l_1$, where hidden layers compute $\mathbf{x}^{(i)} = \phi(\mathbf{W}^{(i)} \mathbf{x}^{(i-1)} + \mathbf{b}^{(i)})$ with activation $\phi$ being the ReLU activation function $\phi(x)=\max(0,x)$.
A quantization configuration $C = \langle \tau, Q, F \rangle$ specifies signedness $\tau \in \{+, \pm\}$, total bits $Q$, and fractional bits $F$, which together define representable bounds $C^{\text{lb}}$ and $C^{\text{ub}}$ (with $C^{\text{lb}}=0$ and $C^{\text{ub}}=2^Q-1$ if $\tau=+$, and $C^{\text{lb}}=-2^{Q-1}$ and $C^{\text{ub}}=2^{Q-1}-1$ otherwise).
Given configurations $C_\text{in}, C_w, C_b$ and $C_h$ for input, weights, biases, and hidden layers, a QNN $\widehat{\mathcal{N}} : \mathbb{Z}^d \to \mathbb{R}^s$ computes:
\begin{align*}
\widehat{\mathbf{a}}^{(i)}
&= 2^{F_i}\mathbf{W}_q^{(i)}
   \widehat{\mathbf{x}}^{(i-1)}
   + 2^{F_h-F_b}\mathbf{b}_q^{(i)},\\
\widehat{\mathbf{x}}^{(i)}
&= \operatorname{clamp}\bigl(
   \lfloor\widehat{\mathbf{a}}^{(i)}\rceil,
   0,C_h^{\mathrm{ub}}\bigr), \quad i<N,\\
\widehat{\mathbf{y}}
&= 2^{-F_h}\widehat{\mathbf{a}}^{(N)}.
\end{align*}
Here, $\mathbf{W}_q^{(i)}$ and $\mathbf{b}_q^{(i)}$ denote
integer-quantized weights and biases, obtained by rounding
and clamping the corresponding tensors scaled by
$2^{F_w}$ and $2^{F_b}$, respectively.
The integer input $\widehat{\mathbf{x}}^{(0)}$ is obtained
similarly using $F_{\mathrm{in}}$.
We set $F_1=F_h-F_w-F_{\mathrm{in}}$ and
$F_i=-F_w$ for $i>1$.

\subsection{Abstract Interpretation}

Abstract interpretation provides a framework for soundly overapproximating the behavior of complex systems using simpler geometric representations. The main idea is to replace exact set computations, which are usually not computable with operations on abstract domains like boxes or zonotopes that overapproximate the true reachable sets. Each operation on the neural network (affine transformations, activation functions) is replaced by an \emph{abstract transformer} that operates on these geometric shapes while guaranteeing that all true outputs are contained within the abstract result. This enables scalable verification: if the abstract overapproximation satisfies a property, the true network must also satisfy it. More formally:

\textbf{Abstract transformers.} For function $f : \mathbb{R}^m \to \mathbb{R}^n$, concrete transformer $T_f : \mathcal{P}(\mathbb{R}^m) \to \mathcal{P}(\mathbb{R}^n)$ maps $T_f(X) = \{f(\mathbf{x}) \mid \mathbf{x} \in X\}$. Abstract transformer $T_f^\# : \mathcal{A}^m \to \mathcal{A}^n$ on domain $\mathcal{A}$ must be sound: $T_f(\gamma^m(a)) \subseteq \gamma^n(T_f^\#(a))$ for all $a \in \mathcal{A}^m$, where $\gamma$ is concretization.

\textbf{Zonotopes.} A zonotope $Z \subseteq \mathbb{R}^n$ is represented as $z(\boldsymbol{\epsilon}) = \mathbf{M} \boldsymbol{\epsilon} + \mathbf{v}$ for $\boldsymbol{\epsilon} \in [-1,1]^m$, where $\mathbf{v} \in \mathbb{R}^n$ is the center and $\mathbf{M} \in \mathbb{R}^{n \times m}$ defines the generators. We also denote this as $Z = \langle \mathbf{M}, \mathbf{v}\rangle$. Their linear structure enables efficient propagation through affine transformations while preserving error dependencies.

\textbf{Constrained zonotopes.} Originally introduced by Scott et al.~\cite{scott2016constrained} with equality constraints, constrained zonotopes have since been extended to inequality constraints in neural network verification~\cite{Lemesle2024}. We use the inequality formulation: $Z^{C} = \{z(\boldsymbol{\epsilon}) \mid \boldsymbol{\epsilon} \in [-1,1]^m, \mathbf{A}\boldsymbol{\epsilon} + \mathbf{b} \geq \mathbf{0}\}$, where $\mathbf{A} \in \mathbb{R}^{K \times m}$, $\mathbf{b} \in \mathbb{R}^K$, which we also represent as $Z^C = (\langle \mathbf{M}, \mathbf{v}\rangle, \mathbf{A}, \mathbf{b})$. This provides tighter over-approximations at increased computational cost (requires LP solvers).

\textbf{Abstract transformers for unquantized networks.} We first present abstract transformers for standard (unquantized) neural network layers, following the established framework~\cite{Gehr2018,singh2018fast}. For an affine layer computing $\mathbf{y} = \mathbf{W}\mathbf{x} + \mathbf{b}$, the abstract transformer is exact for zonotopes: given input zonotope $Z = \langle \mathbf{M}, \mathbf{v}\rangle$, the output is $Z' = \langle \mathbf{W}\mathbf{M}, \mathbf{W}\mathbf{v} + \mathbf{b}\rangle$. For constrained zonotopes, the constraints are preserved: $Z'^C = (\langle \mathbf{W}\mathbf{M}, \mathbf{W}\mathbf{v} + \mathbf{b}\rangle, \mathbf{A}, \mathbf{b})$.

For the ReLU activation $\mathrm{ReLU}(x) = \max(x,0)$, we apply the standard zonotopic abstraction~\cite{Gehr2018} to each neuron independently. Consider a single neuron with input $x$ represented by zonotope $z(\boldsymbol{\epsilon}) = \sum_{j=1}^{m} \mathbf{M}_j \boldsymbol{\epsilon}_j + \mathbf{v}$. The pre-activation bounds are computed as:
\[
l = \mathbf{v} - \sum_{j=1}^{m} |\mathbf{M}_j|, \qquad
u = \mathbf{v} + \sum_{j=1}^{m} |\mathbf{M}_j|.
\]
Three cases are distinguished:
(1) Fully negative ($u \le 0$): $y = 0$;
(2) Fully positive ($l \ge 0$): $y = x$;
(3) Unstable ($l < 0 < u$): a linear relaxation is applied.

For the unstable case, we introduce relaxation parameters
\[
\lambda = \frac{u}{u - l}, \qquad
\mu = -\frac{u l}{2(u - l)}.
\]
For standard zonotopes, we add a fresh noise symbol $\boldsymbol{\epsilon}_{m+1}$ and define:
\[
y = \lambda \left(\sum_{j=1}^{m} \mathbf{M}_j \boldsymbol{\epsilon}_j + \mathbf{v}\right) + \mu \boldsymbol{\epsilon}_{m+1} + \mu.
\]
This abstraction is applied independently to each neuron in each layer.

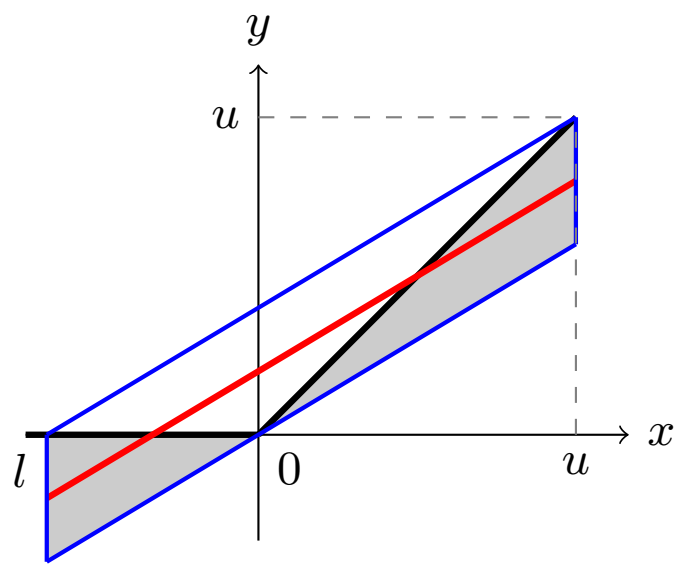
\begin{figure}[t]
\centering
\begin{tikzpicture}[scale=0.7, every node/.style={font=\small}]
    \draw[->] (-2.2, 0) -- (3.5, 0) node[right] {$x$};
    \draw[->] (0, -1) -- (0, 3.5) node[above] {$y$};
    
    \fill[gray!40] (0, 0) -- (3, 1.8) -- (3, 3) -- cycle;
    \fill[gray!40] (0, 0) -- (-2, 0) -- (-2, -1.2) -- cycle;
    
    \draw[very thick, black] (-2.2, 0) -- (0, 0);
    \draw[very thick, black] (0, 0) -- (3, 3);
    
    \draw[very thick, red] (-2, -0.6) -- (3, 2.4);

    \draw[thick, blue] (-2, -1.2) -- (3, 1.8);  
    \draw[thick, blue] (-2, 0) -- (3, 3);  
    \draw[thick, blue] (-2, -1.2) -- (-2, 0);
    \draw[thick, blue] (3, 3) -- (3, 1.8);
    
    \draw[dashed, gray] (0, 3) -- (3, 3);
    \draw[dashed, gray] (3, 0) -- (3, 3);

    \node[below left] at (-2, 0) {$l$};
    \node[below right] at (0, 0) {$0$};
    \node[below] at (3, 0) {$u$};
    \node[left] at (0, 3) {$u$};
\end{tikzpicture}
\caption{Abstract ReLU for the unstable case. The affine abstraction 
$y \in [\lambda x, \lambda x + 2\mu]$ (central affine line with upper and lower error bounds, shown in red and blue) soundly over-approximates the ReLU graph (solid black), with the grey regions eliminated by the two added constraints to tighten the over-approximation.}
\label{fig:relu}
\end{figure}

For constrained zonotopes, following~\cite{Lemesle2024}, the same relaxation is encoded by adding two linear constraints per unstable neuron, requiring the abstract output $y = \lambda(\sum_{j} \mathbf{M}_j \boldsymbol{\epsilon}_j + \mathbf{v}) + \mu \boldsymbol{\epsilon}_{m+1} + \mu$ to satisfy $y \geq 0$ and $y \geq x$. Expanding these conditions and writing the constrained zonotope as $Z^C = (\langle \mathbf{M}, \mathbf{v}\rangle, \mathbf{A}, \mathbf{b})$, the constraints are appended to $\mathbf{A} \in \mathbb{R}^{K \times m}$ and $\mathbf{b} \in \mathbb{R}^{K}$ as:
\[
\mathbf{A}_{K+1,:} = (\lambda\mathbf{M}, \mu), \quad \mathbf{b}_{K+1} = \lambda \mathbf{v} + \mu 
\]
\[
\mathbf{A}_{K+2,:} = ((\lambda-1)\mathbf{M}, \mu), \quad \mathbf{b}_{K+2} = (\lambda - 1) \mathbf{v} + \mu
\]
These eliminate infeasible ReLU outputs while preserving soundness, as illustrated in Figure~\ref{fig:relu}, and $\mathbf{A}$ is padded with a zero column before the rows are appended. The process is repeated for each unstable neuron across all layers.

\section{Abstract Transformers for Quantized Networks}
\label{sec:round-clamp}

To compute quantization error bounds, we propagate abstract representations through both the original network and its quantized counterpart, then compute the difference between their outputs. This requires abstract transformers for standard operations (affine layers, ReLU) in the original network, as well as novel transformers for quantization-specific operations (rounding, clamping) that appear in the quantized network.

We now present a central contribution to our work: novel abstract transformers for rounding and clamping operations that handle the discrete nature of quantization while maintaining sound overapproximations. We build upon the framework of~\cite{Lemesle2024} to design transformers for both standard and constrained zonotopes.

Quantized neural networks apply rounding followed by clamping at each activation: $y_i = \text{clamp}(\lfloor x_i \rceil, 0, C^{\mathrm{ub}})$. A naive approach would abstract these operations separately, first propagating the zonotope through an abstract round transformer, then through an abstract clamp transformer. However, this cascading introduces two problems: (1) each operation adds a fresh noise symbol, causing rapid growth in the zonotope representation, and (2) treating operations independently ignores their compositional structure, leading to loose over-approximations.

We instead derive a \emph{unified} abstraction that captures both operations simultaneously using a single affine transformation $y_i \in [\widehat{\lambda} x_i + \widehat{\mu} - \widehat{e}, \widehat{\lambda} x_i + \widehat{\mu} + \widehat{e}]$. This approach halves the noise symbol count and exploits the geometric structure of the composed operation to achieve significantly tighter bounds.

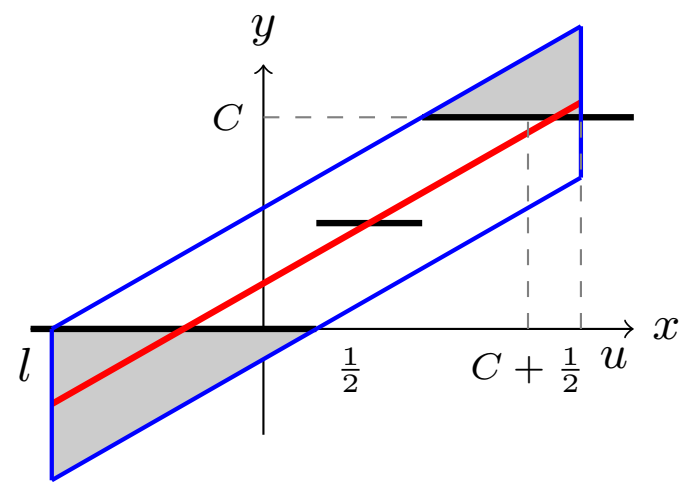
\begin{figure}[t]
\centering
\begin{tikzpicture}[scale=0.7, every node/.style={font=\small}]
    \draw[->] (-2.2, 0) -- (3.5, 0) node[right] {$x$};
    \draw[->] (0, -1) -- (0, 2.5) node[above] {$y$};
    
    \fill[gray!40] (-2, 0) -- (0.5, 0) -- (-2, -1.43) -- cycle;
    \fill[gray!40] (1.5, 2) -- (3, 2) -- (3, 2.86) -- cycle;
    
    \draw[very thick, black] (-2.2, 0) -- (0.5, 0);
    \draw[very thick, black] (0.5, 1) -- (1.5, 1);
    \draw[very thick, black] (1.5, 2) -- (3.5, 2);
    
    \draw[very thick, red] (-2, -0.71) -- (3, 2.14);

    \draw[thick, blue] (-2, -1.43) -- (3, 1.43);  
    \draw[thick, blue] (-2, 0) -- (3, 2.86);  
    \draw[thick, blue] (-2, -1.43) -- (-2, 0);
    \draw[thick, blue] (3, 2.86) -- (3, 1.43);
    
    \draw[dashed, gray] (3, 0) -- (3, 2);
    \draw[dashed, gray] (2.5, 0) -- (2.5, 2);
    \draw[dashed, gray] (0, 2) -- (1.5, 2);
    
    \node[below left] at (-2, 0) {$l$};
    \node[below right] at (0.5, 0) {\scriptsize$\frac{1}{2}$};
    \node[below] at (2.5, 0) {\scriptsize$C + \frac{1}{2}$};
    \node[below right] at (3, 0) {$u$};
    \node[left] at (0, 2) {\scriptsize$C$};
\end{tikzpicture}

\caption{Abstract round-clamp for Case 4: input range $[l, u]$ spans both rounding and clamping thresholds. The affine abstraction $y \in [\widehat{\lambda} x + \widehat{\mu} - \widehat{e}, \widehat{\lambda} x + \widehat{\mu} + \widehat{e}]$ (central affine line with upper and lower error bounds) soundly over-approximates the stepped function (solid black), with the grey regions eliminated by the two box constraints $y \geq 0$ and $y \leq C$. The complete formulation (Appendix~\ref{app:round_clamp}) adds a third, oblique constraint.}
\label{fig:round_clamp_case4}
\end{figure}

The round-clamp function exhibits a characteristic stepped structure (see Figure~\ref{fig:round_clamp_case4}): flat regions where outputs are clamped, and an increasing step function with jumps at rounding thresholds within the clamping bounds. The optimal affine abstraction depends critically on how the input range $[l_i, u_i]$ intersects these regions, leading to six exhaustive cases based on the positions of the lower threshold $-1/2$ and clamp threshold $C^{\mathrm{ub}} + 1/2$ relative to $[l_i, u_i]$.

Throughout, $[l, u]$ denotes the pre-activation interval of neuron $i$, $r_l = \lfloor l \rceil$, $r_u = \min(\lfloor u \rceil,C)$ with $C = C^{\mathrm{ub}}$. Here, $\lfloor x\rceil$ denotes rounding to the
nearest integer, with ties broken toward the even integer.

\textbf{Cases 1-2} (trivial): If $u \leq -1/2$ the output is clamped to $0$
($\widehat{\lambda}=0$, $\widehat{\mu}=0$); if $l \geq C+1/2$ it is clamped to
$C$ ($\widehat{\lambda}=0$, $\widehat{\mu}=C$). No error term or constraint is
needed in either case.

\textbf{Case 3} ($-\tfrac12 \leq l < C+\tfrac12$, $-\tfrac12 < u \leq C+\tfrac12$): Rounding is abstracted within the clamp range,
including upper clamping at $u=C+\tfrac12$. If $r_l = r_u$ the output is the constant $r_l$ with no abstraction error
(Case~3a). Otherwise, at least one rounding threshold is crossed and we use
the identity slope with a $\pm 1/2$ error term:
\begin{equation*}
\widehat{\lambda} = 1,\quad \widehat{\mu} = 0,\quad \widehat{e} = 0.5.
\end{equation*}
Two constraints $y \geq r_l$ and $y \leq r_u$ tighten the output to the
discrete range $[r_l, r_u]$ which excludes outputs outside the reachable range (Case~3b).

\textbf{Case 4} ($l \leq -\tfrac12$,
$u \geq C+\tfrac12$,
$(l,u)\neq(-\tfrac12,C+\tfrac12)$): The range spans both clamp
thresholds. To minimise abstraction error, we select among two candidate
slopes, each tight at one extreme:
\begin{equation*}
\widehat{\lambda}_1 = \frac{C}{u - 0.5}, \qquad
\widehat{\lambda}_2 = \frac{C}{C - l - 0.5},
\end{equation*}
choosing $\widehat{\lambda} = \widehat{\lambda}_1$ when $C - l \leq u$ and
$\widehat{\lambda}_2$ otherwise. The offset and error coefficient are set to
centre the enclosing strip:
\begin{align*}
\widehat{\mu}_1 &= \max(-\widehat{\lambda}\,l,\;
    C - \widehat{\lambda}(C-0.5)), \\
\widehat{\mu}_2 &= \min(-0.5\,\widehat{\lambda},\;
    C - \widehat{\lambda}\,u),\\
\widehat{\mu} &= \tfrac{1}{2}(\widehat{\mu}_1+\widehat{\mu}_2),\qquad
\widehat{e} = \tfrac{1}{2}(\widehat{\mu}_1-\widehat{\mu}_2).
\end{align*}
Two constraints $y \geq 0$ and $y \leq C$ eliminate the portions of the
affine strip that violate the clamp.

\textbf{Case 5} ($l < -\tfrac12$,
$-\tfrac12 < u < C+\tfrac12$): The range crosses only the
lower threshold. The slope is chosen so that the upper edge of the strip is tight at $(l,0)$ and $(r_u-\tfrac12,r_u)$:
\begin{equation*}
\widehat{\lambda} = \frac{r_u}{r_u - l - 0.5},\quad
\widehat{\mu}_1 = -\widehat{\lambda}\,l,\quad
\widehat{\mu}_2 = -\frac{\widehat{\lambda}}{2},
\end{equation*}
with $\widehat{\mu}$ and $\widehat{e}$ as above. Two constraints $y \geq 0$
and $y \leq r_u$ are added.

\textbf{Case 6} ($-\tfrac12 < l < C+\tfrac12$,
$u > C+\tfrac12$): Symmetric to Case~5 with
the roles of the lower and upper thresholds exchanged. The slope is
$\widehat{\lambda} = (C - r_l)/(u - r_l - 0.5)$, and two constraints
$y \geq r_l$ and $y \leq C$ are added.

In all non-trivial cases the zonotope update takes the form
$\mathbf{v}_i^{\mathrm{out}} = \widehat{\lambda}\,\mathbf{v}_i + \widehat{\mu}$ and
$\mathbf{M}_{i,:}^{\mathrm{out}} = (\widehat{\lambda}\,\mathbf{M}_{i,:} \;\|\; \widehat{e})$, by
appending a fresh noise generator of magnitude $\widehat{e}$.


The zonotopic and constrained zonotopic formulations are
provided in full in Appendix~\ref{app:round_clamp}. We provide the \textit{complete}
variant of the constrained abstract round-clamp, from which the main,
lighter formulation follows by retaining only the first two constraints
per neuron. This complete variant incorporates oblique
convex-hull constraints in Cases~3b, 4, 5, and~6, and diagonal rounding
bounds ($y \geq x - \tfrac{1}{2}$ and $y \leq x + \tfrac{1}{2}$) in
Cases~5 and~6, yielding up to four constraints per neuron.

\begin{theorem}[Soundness and optimal tightness]
\label{thm:roundclamp_main}
For every case, the round-clamp abstract transformer is sound: for all
$x \in [l, u]$,
\[
\text{clamp}(\lfloor x \rceil, 0, C) \in
[\widehat{\lambda}\,x + \widehat{\mu} - \widehat{e},\;
 \widehat{\lambda}\,x + \widehat{\mu} + \widehat{e}].
\]
Moreover, the parameters
$(\widehat{\lambda},\widehat{\mu},\widehat{e})$
minimise the width of the enclosing affine strip. The theoretical complete constrained formulation is convex-hull optimal under the conditions established in Appendix~\ref{app:constrained_soundness}.
\end{theorem}

Soundness and affine-strip optimality are established in
Appendix~\ref{app:soundness_optimality}, while convex-hull
optimality of the complete constrained formulation is proved
in Appendix~\ref{app:constrained_soundness}. We adopt the light variant (two box constraints
per unstable neuron) as our main method: our ablation study (Ablation~1) shows it is negligibly looser than the complete constrained formulation (less than $0.6\%$ difference in output interval width) while being 35-46\% faster, which makes it the most practical choice.

\section{Quantization error zonotopic representations}

We now aim to represent quantization errors using zonotopes (and constrained
zonotopes). By concretizing the zonotopes, we can evaluate the fidelity of
quantization on a network and potentially establish upper bounds beyond which
quantization is no longer acceptable.

\subsection{Defining the quantization error zonotope}

We now describe how to compute formal bounds on the quantization error between
an original network $\mathcal{N}$ and its quantized counterpart
$\widehat{\mathcal{N}}$. Our approach consists in propagating the zonotopic
abstractions through the idealized (unquantized) and quantized networks
independently and taking their difference, which is then concretized to obtain
quantization error interval bounds. Specifically, we represent the quantized
network's zonotope as $\widehat{\mathbf{z}}$, the idealized network's zonotope
as $\mathbf{z}$, and compute the quantization error zonotope as a function of the generator and centers of $\widehat{\mathbf{z}}$ and $\mathbf{z}$.
\textbf{Setup.} Let $\mathbf{x}_{0} \in X \subset \mathbb{R}^{d}$ be an input
point and $R > 0$ a radius such that the input region is the box
$B = D_{R}^{\|\cdot\|_{\infty}}(\mathbf{x}_{0}) \subset X$. Let
$\mathcal{N} = l_{N} \circ \dots \circ l_{1}$ be the original (idealized)
network with real-valued parameters, and $\widehat{\mathcal{N}}$ its quantized
counterpart with a specific quantization configuration (fractional bits $F_w, F_b$
and clamp bounds $C_w^{lb}, C_w^{ub}, C_b^{lb}, C_b^{ub}$).

\textbf{Initial zonotopes.} We define input zonotopes for both networks:
\begin{itemize}
    \item For the idealized network: center $\mathbf{v}^{(0)} = \mathbf{x}_{0}$
    and generator matrix $\mathbf{M}^{(0)} = R \cdot \mathbf{I}_{d}$, giving
    $\mathbf{z}^{(0)}(\boldsymbol{\epsilon}^{(0)}) =
    \mathbf{M}^{(0)}\boldsymbol{\epsilon}^{(0)} + \mathbf{v}^{(0)}$,
    where $\boldsymbol{\epsilon}^{(0)} \in [-1,1]^d$.
    \item For the quantized network: $\widehat{\mathbf{M}}^{(0)} \in
    \mathbb{R}^{d \times d}$ and $\widehat{\mathbf{v}}^{(0)} \in \mathbb{R}^d$
    are obtained by quantizing the vertices of $B$, computing the center of the resulting quantized box, and extracting per-dimension radii, giving
    $\widehat{\mathbf{z}}^{(0)}(\widehat{\boldsymbol{\epsilon}}^{(0)}) =
    \widehat{\mathbf{M}}^{(0)}\widehat{\boldsymbol{\epsilon}}^{(0)} +
    \widehat{\mathbf{v}}^{(0)}$.
\end{itemize}

\textbf{Remark.} We do not use a shared symbol for $\boldsymbol{\epsilon}^{(0)}$ and $\widehat{\boldsymbol{\epsilon}}^{(0)}$, since applying the round-clamp zonotope transformer directly to $\mathbf{z}^{(0)}$ yields slightly looser bounds on average, particularly at small perturbation radii (Ablation~2).

\textbf{Quantized weights and biases.} For each layer
$i \in \{1, \dots, N\}$ with weight matrix
$\mathbf{W}^{(i)} \in \mathbb{R}^{n_i \times n_{i-1}}$ and bias vector
$\mathbf{b}^{(i)} \in \mathbb{R}^{n_i}$, the quantized parameters are:
\begin{itemize}
    \item $\widehat{\mathbf{W}}^{(i)} = 2^{-F_w}
    \text{clamp}\!\left(\lfloor 2^{F_w} \mathbf{W}^{(i)} \rceil,
    C_{w}^{lb}, C_{w}^{ub}\right) \in \mathbb{R}^{n_i \times n_{i-1}}$,
    \item $\widehat{\mathbf{b}}^{(i)} = 2^{-F_b}
    \text{clamp}\!\left(\lfloor 2^{F_b} \mathbf{b}^{(i)} \rceil,
    C_{b}^{lb}, C_{b}^{ub}\right) \in \mathbb{R}^{n_i}$.
\end{itemize}

In the abstract propagation, the quantized input is represented
in dequantized units, whereas hidden activations are represented
in integer units. We therefore use the effective affine parameters
\[
\widehat{\mathbf{W}}_{\mathrm{eff}}^{(i)}
=
\begin{cases}
2^{F_h}\widehat{\mathbf{W}}^{(i)}, & i=1,\\
\widehat{\mathbf{W}}^{(i)}, & i>1,
\end{cases}
\qquad
\widehat{\mathbf{b}}_{\mathrm{eff}}^{(i)}
=2^{F_h}\widehat{\mathbf{b}}^{(i)}.
\]
The idealized network is propagated analogously in
$2^{F_h}$-scaled coordinates, which preserves its
behavior by the positive homogeneity of ReLU.

\textbf{Layer-wise propagation.} At each layer $i$, both networks receive as
input the post-activation zonotopes from the previous layer:
\begin{itemize}
    \item $\mathbf{z}_{a}^{(i-1)}(\boldsymbol{\epsilon}_{a}^{(i-1)}) =
    \mathbf{M}_{a}^{(i-1)} \boldsymbol{\epsilon}_{a}^{(i-1)} +
    \mathbf{v}_{a}^{(i-1)}$: idealized post-activation zonotope, with
    $\mathbf{M}_{a}^{(i-1)} \in \mathbb{R}^{n_{i-1} \times m_{i-1}}$,
    $\mathbf{v}_{a}^{(i-1)} \in \mathbb{R}^{n_{i-1}}$,
    \item $\widehat{\mathbf{z}}_{a}^{(i-1)}(\widehat{\boldsymbol{\epsilon}}_{a}^{(i-1)})
    = \widehat{\mathbf{M}}_{a}^{(i-1)} \widehat{\boldsymbol{\epsilon}}_{a}^{(i-1)} +
    \widehat{\mathbf{v}}_{a}^{(i-1)}$: quantized post-activation zonotope,
    with $\widehat{\mathbf{M}}_{a}^{(i-1)} \in \mathbb{R}^{n_{i-1} \times
    m_{i-1}}$, $\widehat{\mathbf{v}}_{a}^{(i-1)} \in \mathbb{R}^{n_{i-1}}$,
    \item $\widehat{\boldsymbol{\epsilon}}_{a}^{(i-1)} \in [-1,1]^{m_{i-1}}$:
    accumulated noise symbols up to layer $(i-1)$, that is, $m_0 = d$, $m_i = m_{i-1} + n_i$ for $i \in \{1,\dots, N-1\}$, and $m_N = m_{N-1}$, since the final affine layer adds no noise symbols.
\end{itemize}

\textbf{Affine transformation.} Applying the quantized affine
transformation to $\widehat{\mathbf{z}}_{a}^{(i-1)}$:
\begin{align*}
    \widehat{\mathbf{z}}^{(i)}(\widehat{\boldsymbol{\epsilon}}^{(i)})
    &= \widehat{\mathbf{W}}_{\mathrm{eff}}^{(i)}
       \widehat{\mathbf{z}}_{a}^{(i-1)}(\widehat{\boldsymbol{\epsilon}}_{a}^{(i-1)})
       + \widehat{\mathbf{b}}_{\mathrm{eff}}^{(i)} \\
    &= \bigl(\widehat{\mathbf{W}}_{\mathrm{eff}}^{(i)}\widehat{\mathbf{M}}_{a}^{(i-1)}\bigr)
       \widehat{\boldsymbol{\epsilon}}_{a}^{(i-1)}
       + \bigl(\widehat{\mathbf{W}}_{\mathrm{eff}}^{(i)}\widehat{\mathbf{v}}_{a}^{(i-1)}
       + \widehat{\mathbf{b}}_{\mathrm{eff}}^{(i)}\bigr) \\
    &= \widehat{\mathbf{M}}^{(i)} \widehat{\boldsymbol{\epsilon}}^{(i)}
      + \widehat{\mathbf{v}}^{(i)},
\end{align*}
with $\widehat{\boldsymbol{\epsilon}}^{(i)} =
\widehat{\boldsymbol{\epsilon}}_{a}^{(i-1)}$. The idealized path is
identical with the hat removed from every quantity, yielding
$\mathbf{z}^{(i)}(\boldsymbol{\epsilon}^{(i)}) =
\mathbf{M}^{(i)} \boldsymbol{\epsilon}^{(i)} + \mathbf{v}^{(i)}$
with $\boldsymbol{\epsilon}^{(i)} = \boldsymbol{\epsilon}_{a}^{(i-1)}$.

\textbf{Activation transformation.} After the affine layer, we apply activation functions: the ReLU for the idealized network, round-clamp for the quantized network, which are abstracted via the
linear relaxations parameterized by:
\begin{itemize}
    \item $\boldsymbol{\lambda}^{(i)},\,\widehat{\boldsymbol{\lambda}}^{(i)}
    \in \mathbb{R}^{n_i \times n_i}$: diagonal slope matrices,
    \item $\boldsymbol{\mu}^{(i)},\,\widehat{\boldsymbol{\mu}}^{(i)}
    \in \mathbb{R}^{n_i}$: offset vectors,
    \item $\mathbf{E}^{(i)},\,\widehat{\mathbf{E}}^{(i)}
    \in \mathbb{R}^{n_i \times n_i}$: diagonal matrices of abstraction
    error coefficients,
    \item${\boldsymbol{\varepsilon}}^{(i)},\widehat{{\boldsymbol{\varepsilon}}}^{(i)} \in [-1, 1]^{n_i}$: fresh noise symbols introduced at layer $i$ in the unquantized and the quantized network .
\end{itemize}

The quantized post-activation zonotope is:
\begin{align*}
    \widehat{\mathbf{z}}_{a}^{(i)}(\widehat{\boldsymbol{\epsilon}}_{a}^{(i)})
    &= \widehat{\boldsymbol{\lambda}}^{(i)} \widehat{\mathbf{z}}^{(i)}
       (\widehat{\boldsymbol{\epsilon}}^{(i)})
       + \widehat{\boldsymbol{\mu}}^{(i)}
       + \widehat{\mathbf{E}}^{(i)} \widehat{{\boldsymbol{\varepsilon}}}^{(i)} \\
    &= \bigl(\widehat{\boldsymbol{\lambda}}^{(i)}\widehat{\mathbf{M}}^{(i)}
       \;\|\; \widehat{\mathbf{E}}^{(i)}\bigr)
       \widehat{\boldsymbol{\epsilon}}_{a}^{(i)}
       + \bigl(\widehat{\boldsymbol{\lambda}}^{(i)}\widehat{\mathbf{v}}^{(i)}
       + \widehat{\boldsymbol{\mu}}^{(i)}\bigr) \\
    &= \widehat{\mathbf{M}}_{a}^{(i)} \widehat{\boldsymbol{\epsilon}}_{a}^{(i)}
      + \widehat{\mathbf{v}}_{a}^{(i)},
\end{align*}
where $\widehat{\boldsymbol{\epsilon}}_{a}^{(i)} =
(\widehat{\boldsymbol{\epsilon}}^{(i)\top} \|
\widehat{\boldsymbol{\varepsilon}}^{(i)\top})^\top \in [-1,1]^{m_i}$ and
$\widehat{\mathbf{M}}_{a}^{(i)} \in \mathbb{R}^{n_i \times m_i}$ is
obtained by concatenating the corresponding column blocks. The idealized
post-activation zonotope
$\mathbf{z}_{a}^{(i)}(\boldsymbol{\epsilon}_{a}^{(i)}) =
\mathbf{M}_{a}^{(i)} \boldsymbol{\epsilon}_{a}^{(i)} + \mathbf{v}_{a}^{(i)}$
is obtained identically with the hat removed, using the ReLU relaxation
$(\boldsymbol{\lambda}^{(i)}, \boldsymbol{\mu}^{(i)}, \mathbf{E}^{(i)})$
and noise symbols $\boldsymbol{\epsilon}_{a}^{(i)} =
(\boldsymbol{\epsilon}^{(i)\top} \| \boldsymbol{\varepsilon}^{(i)\top})^\top$.

These transformations are applied across all hidden layers,
followed by the final affine output layer. Both output zonotopes are then rescaled by $2^{-F_h}$
to obtain real-valued output bounds before computing
their difference. Setting
$\boldsymbol{\epsilon} = (\widehat{\boldsymbol{\epsilon}}^{(N)\top} \| \boldsymbol{\epsilon}^{(N)\top})^\top$, we define the quantization error zonotope as:

\begin{align*}
\widetilde{\mathbf{z}}^{(N)}(\boldsymbol{\epsilon})
&= 2^{-F_h}\left(
\widehat{\mathbf{z}}^{(N)}
(\widehat{\boldsymbol{\epsilon}}^{(N)})
-
\mathbf{z}^{(N)}(\boldsymbol{\epsilon}^{(N)})
\right)\\
&= \widetilde{\mathbf{M}}^{(N)}
\boldsymbol{\epsilon}
+\widetilde{\mathbf{v}}^{(N)},
\end{align*}

where
\[
\widetilde{\mathbf{M}}^{(N)}
= 2^{-F_h}\left(
\widehat{\mathbf{M}}^{(N)}
\;\|\;
-\mathbf{M}^{(N)}
\right),
\]
\[
\widetilde{\mathbf{v}}^{(N)}
= 2^{-F_h}\left(
\widehat{\mathbf{v}}^{(N)}
-\mathbf{v}^{(N)}
\right).
\]
We then concretize $\widetilde{\mathbf{z}}^{(N)}$
to obtain interval bounds on the quantization
error per output dimension.

\subsection{Constrained zonotopic representation of quantization error}

We extend the quantization error representation from standard zonotopes to constrained zonotopes by accumulating linear constraints on $\boldsymbol{\epsilon}$ during both propagations. During idealized propagation, each unstable ReLU neuron contributes 2 constraints, collected into $(\mathbf{A}_{\mathrm{dnn}},\,\mathbf{b}_{\mathrm{dnn}})$. During quantized propagation, each unstable round-clamp neuron contributes 2 box constraints, collected into $(\mathbf{A}_{\mathrm{qnn}},\,\mathbf{b}_{\mathrm{qnn}})$.

Since $\widehat{\boldsymbol{\epsilon}}^{(N)}$ and $\boldsymbol{\epsilon}^{(N)}$ are
independent, the two constraint sets cannot act on the same symbol vector. Instead, each is lifted to the full extended space
$\boldsymbol{\epsilon} = (\widehat{\boldsymbol{\epsilon}}^{(N)\top} \| \boldsymbol{\epsilon}^{(N)\top})^\top
\in [-1,1]^{2 m_N}$: the QNN constraints act only on $\widehat{\boldsymbol{\epsilon}}^{(N)}$ (the first $m_N$ components, comprising $\widehat{\boldsymbol{\epsilon}}^{(0)}$
and the round-clamp noise symbols), while the DNN constraints act only on $\boldsymbol{\epsilon}^{(N)}$ (the last $m_N$ components, comprising $\boldsymbol{\epsilon}^{(0)}$ and the ReLU noise symbols). The two input symbol blocks are entirely independent, so the full constraints matrix and vector are expressed as
\begin{equation*}
    \mathbf{A} =
    \begin{pmatrix}
        \mathbf{A}_{\mathrm{qnn}} & \mathbf{0} \\
        \mathbf{0} & \mathbf{A}_{\mathrm{dnn}}
    \end{pmatrix},
    \qquad
    \mathbf{b} =
    \begin{pmatrix} \mathbf{b}_{\mathrm{qnn}} \\ \mathbf{b}_{\mathrm{dnn}} \end{pmatrix}.
\end{equation*}

The quantization error zonotope has center $\widetilde{\mathbf{v}}^{(N)}$ and generator matrix
$\widetilde{\mathbf{M}}^{(N)}$. Denoting the $k$-th component of $\widetilde{\mathbf{v}}^{(N)}$ by $\widetilde{v}_k$ and the $k$-th row of $\widetilde{\mathbf{M}}^{(N)}$ by $\widetilde{\mathbf{m}}_k^\top$, tighter bounds on the quantization error at
output dimension $k$ are obtained by solving:
\begin{equation*}
    \ell_k = \widetilde{v}_k + \min_{\boldsymbol{\epsilon}}\;
    \widetilde{\mathbf{m}}_k^\top \boldsymbol{\epsilon}, \qquad
    u_k = \widetilde{v}_k + \max_{\boldsymbol{\epsilon}}\;
    \widetilde{\mathbf{m}}_k^\top \boldsymbol{\epsilon},
\end{equation*}
subject to $\boldsymbol{\epsilon} \in [-1,1]^{2 m_N}$ and
$\mathbf{A}\,\boldsymbol{\epsilon} + \mathbf{b} \geq \mathbf{0}$.

\section{Implementation}

\textbf{Language and libraries.} We implement our method in Julia 1.12, using LazySets.jl for zonotope arithmetic, JuMP.jl with the HiGHS solver for linear programming, and ModelVerification.jl for loading network models. Experiments are run on an Apple M5 CPU with 16\,GB of RAM under macOS. The implementation and benchmark scripts are available at \url{https://github.com/robertdebbas/zonotopes}.

\textbf{Constraint accumulation.} During the forward pass, each unstable ReLU neuron adds two linear constraints on the noise variables $\boldsymbol{\epsilon}$: $y_i \geq 0$ and $y_i \geq x_i$. Each unstable round-clamp neuron adds two box constraints: a lower bound ($y \geq r_l$ or $y \geq 0$) and an upper bound ($y \leq r_u$ or $y \leq C$), where $r_l = \mathrm{round}(l)$, $r_u = \min(\mathrm{round}(u),C)$, and $C$ is the clamp ceiling. All constraints are collected across layers and solved jointly at the end.

\textbf{Bound computation.} For each of the $s$ output dimensions, we
solve two LPs (one minimization, one maximization) over the $2 m_N$ noise
variables, subject to $\boldsymbol{\epsilon}_i \in [-1,1]$ and the
accumulated constraints $\mathbf{A}_{k,:}\boldsymbol{\epsilon} +
\mathbf{b}_k \geq 0$. We use HiGHS with a 2-second time limit per LP.
If a solve does not complete in time, we fall back to the plain zonotope bound, which is looser but remains sound.

\section{Experiments}
\label{sec:experiments}

We compare our original and constrained zonotopic methods with a baseline: Symbolic DRA propagation~\cite{Zhang2023}, which tracks quantization error bounds using symbolic polyhedral abstractions. We use the official QEBVerif~\cite{Zhang2023} implementation and evaluate its symbolic propagation mode (DRA-Sym).

\subsection{Experimental Setup}

\textbf{Benchmarks.} We evaluate on the ACAS Xu benchmark~\cite{katz2017reluplex}, a standard testbed for neural network verification. The ACAS Xu networks are feed-forward neural networks with ReLU activations used for aircraft collision avoidance. Each one takes five continuous inputs, produces advisories over multiple maneuver options, and consists of six hidden layers with 50 neurons per layer. We select 10 diverse networks spanning families 1-4 ($\mathrm{ACAS}\_i\_j$
for $(i,j)$ in $(1,1)$, $(1,5)$, $(1,9)$, $(2,2)$, $(2,5)$, $(2,9)$, $(3,3)$, $(3,5)$, $(4,1)$, $(4,5)$), to cover different trained network instances while maintaining reasonable computational costs.

\textbf{Quantization Configuration.}
We test across multiple quantization bitwidths: $Q \in \{6,8,10\}$ bits. Following QEBVerif's configuration~\cite{Zhang2023}, we use $C_{\mathrm{in}}=\langle \pm,8,8\rangle$ for the input,
$C_h=\langle +,Q,Q-2\rangle$ for hidden activations,
and $F_w=F_b=\max(2,Q-2)$ fractional bits for weights and biases which matches the quantization scheme in the DRA baseline implementation.

\textbf{Attack Radii.} We evaluate under varying $\ell_\infty$ perturbations using quantized radii $r \in \{3, 6, 13, 19, 26\}$ (corresponding to floating-point $\epsilon = r/256 \in \{0.012, 0.023, 0.051, 0.074, 0.102\}$), following the experimental protocol of~\cite{Zhang2023}. All perturbations are centered around test input $\mathbf{x}_0 = [-0.2, -0.25, -0.5, -0.3, -0.44]^\top$ (QEBVerif input point 4).

\textbf{Evaluation Metrics.} We measure \emph{precision} using $O_{\text{Diff}}$, the average interval width across all output neurons, which quantifies the tightness of quantization error bounds. We measure \emph{speed} by total verification time (propagation and concretization for zonotopes and constrained zonotopes, and the full verification routine including internal LP solving for DRA-Sym). Our benchmark comprises 150 experiments total ($3$ bitwidths $\times$ $5$ radii $\times$ $10$ networks), all of which produce valid results for all three methods.

\subsection{Overall Performance}

Table~\ref{tab:acasxu_results} summarizes average performance across all 150 experiments. Standard zonotopes take an average of $0.002$~s, but give loose bounds ($O_{\text{Diff}} = 361.1$). Constrained zonotopes achieve tighter bounds ($O_{\text{Diff}} = 230.0$) in $0.382$~s, faster than DRA-Sym ($0.564$~s), which attains the best average precision overall ($O_{\text{Diff}} = 145.3$). However, these averages mask a strong regime dependence as shown in Table~\ref{tab:radius_results}. Constrained zonotopes outperform DRA-Sym on \emph{both} precision and speed at small perturbation radii ($r \leq 6$) and remain faster up to $r = 19$. DRA-Sym is both faster and more precise only at the largest radius ($r = 26$).

\begin{table}[t]
\centering
\begin{tabular}{lcc}
\toprule
\textbf{Method} & \textbf{Avg.\ $O_{\text{Diff}}$} & \textbf{Avg.\ Time (s)}\\
\midrule
Original Zonotope    & 361.1                  & 0.002 \\
Constrained Zonotope & 230.0                  & \textbf{0.382} \\
DRA-Sym              & \textbf{145.3}         & 0.564 \\
\bottomrule
\end{tabular}
\caption{Average performance across all 150 ACAS Xu experiments. Bold marks the better of Constrained Zonotope and DRA-Sym; the Original Zonotope is the fastest method.}
\label{tab:acasxu_results}
\end{table}

\begin{table}[t]
\centering
\setlength{\tabcolsep}{4pt}
\begin{tabular}{ccccccc}
\toprule
 & \multicolumn{2}{c}{\textbf{Zonotope}} &
   \multicolumn{2}{c}{\textbf{Constrained}} &
   \multicolumn{2}{c}{\textbf{DRA-Sym}} \\
\cmidrule(lr){2-3}\cmidrule(lr){4-5}\cmidrule(lr){6-7}
$r$ & $O_{\text{Diff}}$ & Time & $O_{\text{Diff}}$ & Time & $O_{\text{Diff}}$ & Time \\
\midrule
3  & 2.3    & 1\,ms & \textbf{1.4}   & \textbf{94\,ms}  & 8.5            & 422\,ms \\
6  & 7.7    & 1\,ms & \textbf{4.7}   & \textbf{170\,ms} & 14.9           & 485\,ms \\
13 & 175.0  & 3\,ms & 111.9          & \textbf{413\,ms} & \textbf{74.0}  & 611\,ms \\
19 & 512.2  & 1\,ms & 326.5          & \textbf{541\,ms} & \textbf{194.4} & 641\,ms \\
26 & 1108.4 & 3\,ms & 705.6          & 691\,ms          & \textbf{435.0} & \textbf{660\,ms} \\
\bottomrule
\end{tabular}
\caption{Per-radius breakdown. Bold marks the better of Constrained and DRA-Sym for each metric; the unconstrained Zonotope is the fastest at every radius. Constrained zonotopes are tighter and faster at $r \leq 6$; DRA-Sym is tighter from $r=13$ and also faster at $r=26$.}
\label{tab:radius_results}
\end{table}

\subsection{Precision vs Speed Tradeoff}

\begin{figure}[t]
\centering
\includegraphics[width=0.45\textwidth]{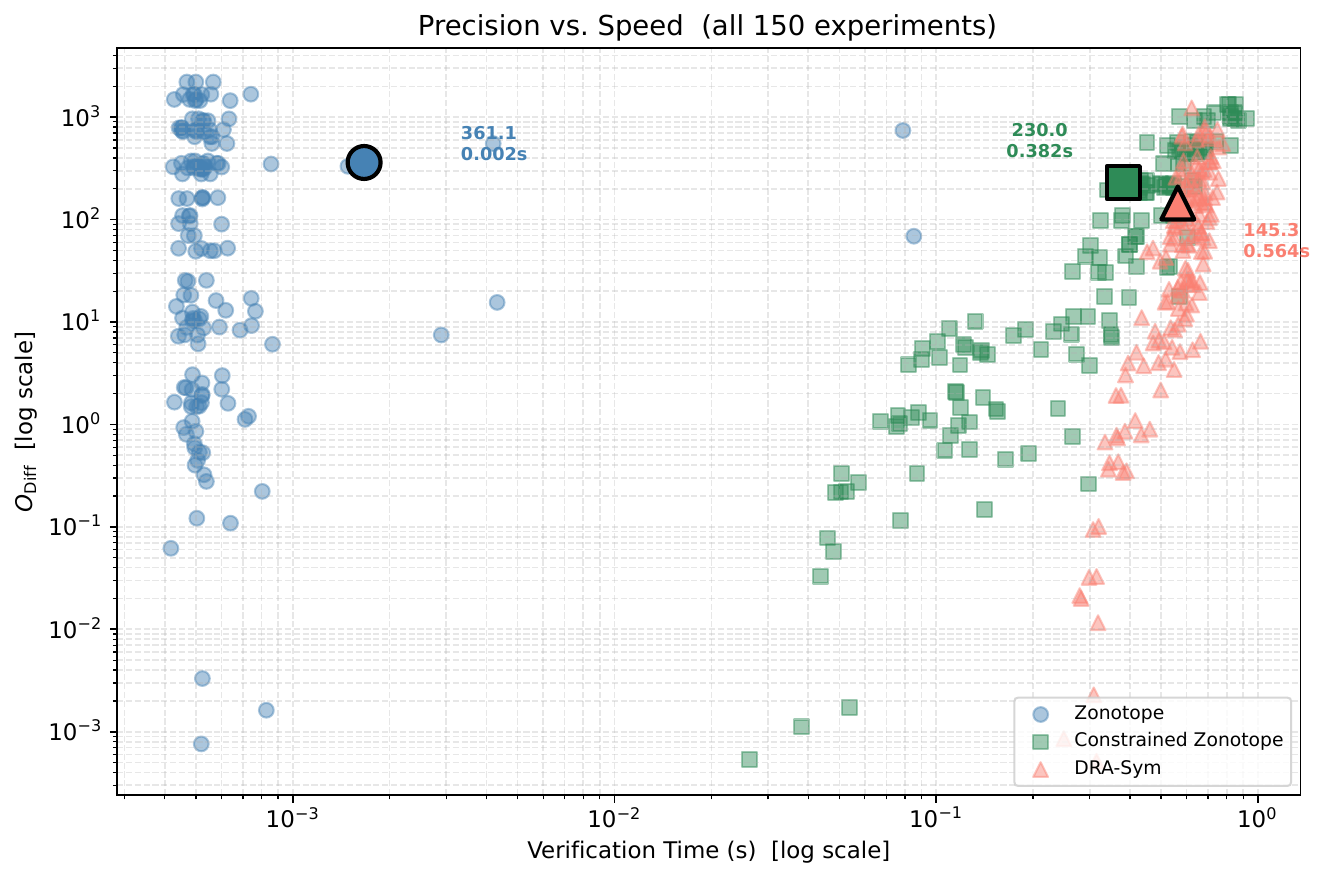}
\caption{Precision vs speed trade-off (log-log scale). Each point is one experiment. At small radii, constrained zonotopes are both faster and more precise than DRA-Sym: runtimes converge only at the largest radius ($r=26$).}
\label{fig:precision_speed}
\end{figure}

Figure~\ref{fig:precision_speed} presents the precision-speed trade-off across all radii and quantization settings. Three distinct clusters are visible. Standard zonotopes form an extremely fast but loose cluster. Constrained zonotopes and DRA-Sym occupy overlapping runtime ranges (${\approx}0.1$-$1$\,s) but differ in precision depending on the perturbation regime: constrained zonotopes
achieve tighter bounds at small radii, while DRA-Sym is more precise at large radii. The two methods reach comparable runtimes only at the largest radius
($r = 26$). Once more, at smaller radii constrained zonotopes are consistently faster.

\subsection{Attack Radius Analysis}
\label{sec:attack_radius}

\begin{figure}[t]
\centering
\includegraphics[width=0.48\textwidth]{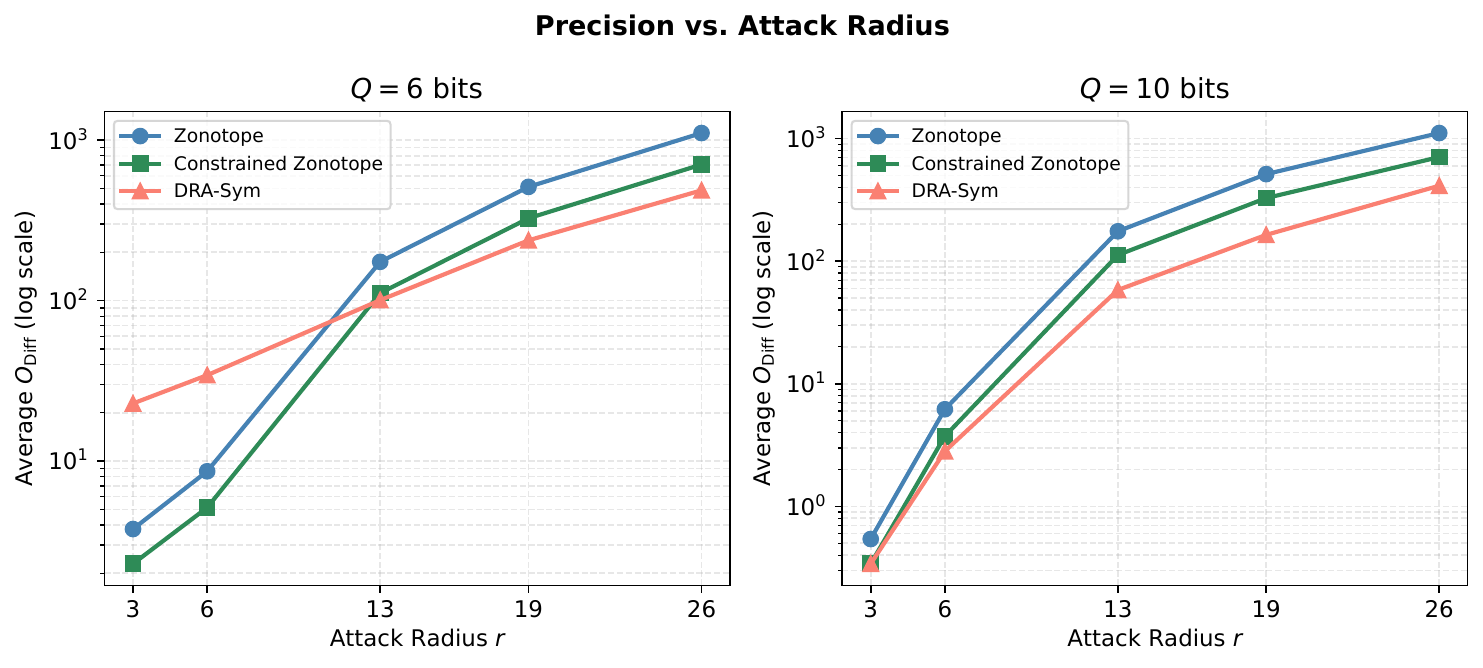}
\caption{Average $O_{\text{Diff}}$ vs.\ attack radius for $Q=6$ and $Q=10$.
Constrained zonotopes are tightest at small radii; DRA-Sym recovers at large
radii.}
\label{fig:odiff_vs_radius}
\end{figure}

Figure~\ref{fig:odiff_vs_radius} shows a clear regime dependence for both bitwidths, though the crossover point differs. At $Q=6$, constrained zonotopes greatly outperform DRA-Sym at small radii ($r \leq 6$) and the crossover occurs at $r=13$. At $Q=10$, the quantization error is small enough that DRA-Sym becomes more precise already at $r=6$, both methods agree very closely at $r=3$. In both cases, the precision advantage of constrained zonotopes is most visible at $Q=6$, where coarser quantization produces larger errors to bound.

\subsection{Quantization Bit-Width Analysis}
\label{sec:bitwidth}

Verification precision also depends on the quantization bit-width. At small radius ($r=6$, $\epsilon=0.023$), constrained zonotopes outperform DRA-Sym at $Q=6$ and $Q=8$ ($O_{\text{Diff}}$ of $5.1$ vs.\ $34.3$ at $Q=6$, and $5.1$ vs.\ $7.6$ at $Q=8$), while at $Q=10$ DRA-Sym is marginally tighter ($2.8$ vs.\ $3.7$). At large radius ($r=19$), DRA-Sym is more precise across all bit-widths. At $r=6$, all methods improve as $Q$ increases, which reflects the smaller quantization error at higher precision. At $r=19$, only DRA-Sym improves with $Q$, while both zonotope bounds vary by less than $0.3\%$ across bit-widths.

\subsection{Ablation 1: Complete vs.\ Main Constrained Formulation}
\label{sec:ablation}

The constrained zonotope framework admits richer constraint sets beyond box
bounds. We compare our main formulation, which considers 2 box constraints per unstable round-clamp neuron, against the complete variant that additionally appends oblique hull edge constraints and diagonal rounding bounds, yielding up to 4 constraints per neuron. The oblique constraints are implemented in division-free form to handle near-degenerate cases without numerical cutoffs. Table~\ref{tab:ablation} reports the per-radius results averaged over all 10 networks and 3 bitwidths.

\begin{table}[t]
\centering
\setlength{\tabcolsep}{5pt}
\begin{tabular}{ccccc}
\toprule
 & \multicolumn{2}{c}{\textbf{Full}} & \multicolumn{2}{c}{\textbf{Main}} \\
\cmidrule(lr){2-3}\cmidrule(lr){4-5}
$r$ & $O_{\text{Diff}}$ & Time & $O_{\text{Diff}}$ & Time \\
\midrule
3  & \textbf{1.40} & 191\,ms & 1.41 & \textbf{104\,ms} \\
6  & \textbf{4.65} & 318\,ms & 4.66 & \textbf{180\,ms} \\
13 & 111.9 & 647\,ms & 111.9 & \textbf{411\,ms} \\
19 & 326.5 & 897\,ms & 326.5 & \textbf{567\,ms} \\
26 & 705.6 & 1098\,ms & 705.6 & \textbf{714\,ms} \\
\bottomrule
\end{tabular}
\caption{Comparison of the full and main constrained formulations across different perturbation radii.}
\label{tab:ablation}
\end{table}

The complete variant provides essentially no precision gain: the largest absolute difference in $O_{\text{Diff}}$ is $0.011$ at $r=6$, less than $0.6\%$ of the bound value. In contrast, the main formulation is consistently faster across all radii, reducing runtime by $43$-$46\%$ at small radii
($r \leq 6$) and by $35$-$37\%$ at larger radii. The overhead of the extra constraints in the full variant is not justified by the negligible improvement in bound tightness which suggests that the box constraints alone capture the essential structure of the round-clamp abstraction.

\subsection{Ablation 2: Minkowski Sum vs.\ Shared Input Abstraction}
\label{sec:ablation-mink}

An alternative to the Minkowski sum construction is
to share the input noise variables between the QNN
and DNN paths, since both are evaluated on the same
underlying input, with the QNN receiving its
quantized representation. In the shared construction, the QNN input is abstracted via the round-clamp
zonotope transformer rather than vertex quantization, allowing the first $d$ generator columns to be subtracted before concatenation. Across all 150 experiments, both constructions achieve
nearly identical average precision ($361.18$ for shared
input and $361.11$ for Minkowski sum). The independent construction yields tighter bounds at small perturbation radii, while the shared construction is marginally tighter at larger radii. We retain independent propagation for consistency with our constrained zonotope formulation.

\section{Conclusion}

The nature of zonotopes, defined as Minkowski sums of generators, allowed us to naturally express quantized executions as perturbations of the idealized execution, represented as the Minkowski sum of a central zonotope and a quantization zonotope that can be decomposed into quantization noise terms. Our experiments highlight constrained zonotopes as a practical choice for quantization error bound verification, offering a favorable balance of precision and efficiency, particularly at small perturbation radii where they significantly outperform the state-of-the-art symbolic DRA approach, while DRA-Sym recovers at larger radii where its fixed-cost symbolic propagation becomes competitive. These complementary behaviors suggest a hybrid method selection strategy, with standard zonotopes remaining attractive for large models or rapid screening.

Finally, our comparison focuses on single-pass abstract interpretation. While the full QEBVerif framework achieves high precision by combining DRA with MILP refinement, this comes at the cost of significantly increased verification time. Standard and constrained zonotopes provide a robust foundation for similar approaches and could potentially benefit from additional techniques, for instance through hybrid zonotopes~\cite{Bird2021,Ortiz2022}, which exactly represent piecewise-linear functions such as ReLU activations using binary MILP variables.

\bibliography{references}

\begin{thebibliography}{18}
\providecommand{\natexlab}[1]{#1}

\bibitem[{Amir et~al.(2021)Amir, Wu, Barrett, and Katz}]{amir2021smt}
Amir, G.; Wu, H.; Barrett, C.; and Katz, G. 2021.
\newblock An {SMT}-based approach for verifying binarized neural networks.
\newblock In \emph{International Conference on Tools and Algorithms for the Construction and Analysis of Systems}, 203--222. Springer.

\bibitem[{Bird et~al.(2023)Bird, Pangborn, Jain, and Koeln}]{Bird2021}
Bird, T.~J.; Pangborn, H.~C.; Jain, N.; and Koeln, J.~P. 2023.
\newblock Hybrid Zonotopes: A New Set Representation for Reachability Analysis of Mixed Logical Dynamical Systems.
\newblock \emph{Automatica}, 154: 111107.

\bibitem[{Gehr et~al.(2018)Gehr, Mirman, Drachsler-Cohen, Tsankov, Chaudhuri, and Vechev}]{Gehr2018}
Gehr, T.; Mirman, M.; Drachsler-Cohen, D.; Tsankov, P.; Chaudhuri, S.; and Vechev, M.~T. 2018.
\newblock AI2: Safety and Robustness Certification of Neural Networks with Abstract Interpretation.
\newblock In \emph{2018 IEEE Symposium on Security and Privacy (SP)}, 3--18.

\bibitem[{Goubault and Putot(2011)}]{vmcai11}
Goubault, E.; and Putot, S. 2011.
\newblock Static analysis of finite precision computations.
\newblock In \emph{Proceedings of the 12th International Conference on Verification, Model Checking, and Abstract Interpretation}, VMCAI'11, 232–247. Berlin, Heidelberg: Springer-Verlag.
\newblock ISBN 9783642182747.

\bibitem[{Henzinger, Lechner, and {\v{Z}}ikeli{\'c}(2021)}]{henzinger2021scalable}
Henzinger, T.~A.; Lechner, M.; and {\v{Z}}ikeli{\'c}, D. 2021.
\newblock Scalable verification of quantized neural networks.
\newblock In \emph{Proceedings of the AAAI Conference on Artificial Intelligence}, volume~35, 3787--3795.

\bibitem[{Huang et~al.(2024)Huang, Wu, Yang, Daukantas, Wu, Zhang, and Barrett}]{huang2024efficient}
Huang, P.; Wu, H.; Yang, Y.; Daukantas, I.; Wu, M.; Zhang, Y.; and Barrett, C. 2024.
\newblock Towards Efficient Verification of Quantized Neural Networks.
\newblock In \emph{Proceedings of the AAAI Conference on Artificial Intelligence}, volume~38, 21152--21160.

\bibitem[{Huang et~al.(2017)Huang, Kwiatkowska, Wang, and Wu}]{huang2017safety}
Huang, X.; Kwiatkowska, M.; Wang, S.; and Wu, M. 2017.
\newblock Safety verification of deep neural networks.
\newblock In \emph{International Conference on Computer Aided Verification}, 3--29. Springer.

\bibitem[{Katz et~al.(2017)Katz, Barrett, Dill, Julian, and Kochenderfer}]{katz2017reluplex}
Katz, G.; Barrett, C.; Dill, D.~L.; Julian, K.; and Kochenderfer, M.~J. 2017.
\newblock Reluplex: An efficient {SMT} solver for verifying deep neural networks.
\newblock In \emph{International Conference on Computer Aided Verification}, 97--117. Springer.

\bibitem[{Lemesle, Lehmann, and Gall(2024)}]{Lemesle2024}
Lemesle, A.; Lehmann, J.; and Gall, T.~L. 2024.
\newblock Neural Network Verification with PyRAT.
\newblock arXiv:2410.23903.

\bibitem[{Liu et~al.(2021)Liu, Arnon, Lazarus, Strong, Barrett, and Kochenderfer}]{liu2021survey}
Liu, C.; Arnon, T.; Lazarus, C.; Strong, C.; Barrett, C.; and Kochenderfer, M.~J. 2021.
\newblock Algorithms for verifying deep neural networks.
\newblock \emph{Foundations and Trends in Optimization}, 4(3-4): 244--404.

\bibitem[{Nagel et~al.(2021)Nagel, Fournarakis, Amjad, Bondarenko, van Baalen, and Blankevoort}]{nagel2021whitepaperneuralnetwork}
Nagel, M.; Fournarakis, M.; Amjad, R.~A.; Bondarenko, Y.; van Baalen, M.; and Blankevoort, T. 2021.
\newblock A White Paper on Neural Network Quantization.
\newblock arXiv:2106.08295.

\bibitem[{Ortiz et~al.(2023)Ortiz, Vellucci, Koeln, and Ruths}]{Ortiz2022}
Ortiz, J.; Vellucci, A.; Koeln, J.; and Ruths, J. 2023.
\newblock Hybrid Zonotopes Exactly Represent ReLU Neural Networks.
\newblock \emph{arXiv preprint arXiv:2304.02755}.

\bibitem[{Scott et~al.(2016)Scott, Raimondo, Marseglia, and Braatz}]{scott2016constrained}
Scott, J.~K.; Raimondo, D.~M.; Marseglia, G.~R.; and Braatz, R.~D. 2016.
\newblock Constrained zonotopes: A new tool for set-based estimation and fault detection.
\newblock \emph{Automatica}, 69: 126--136.

\bibitem[{Singh et~al.(2018)Singh, Gehr, Mirman, P{\"u}schel, and Vechev}]{singh2018fast}
Singh, G.; Gehr, T.; Mirman, M.; P{\"u}schel, M.; and Vechev, M. 2018.
\newblock Fast and effective robustness certification.
\newblock In \emph{Advances in Neural Information Processing Systems}, volume~31, 10802--10813.

\bibitem[{Song et~al.(2023)Song, Sun, Mustafa, and Cordeiro}]{song2023qnnrepairquantizedneuralnetwork}
Song, X.; Sun, Y.; Mustafa, M.~A.; and Cordeiro, L.~C. 2023.
\newblock QNNRepair: Quantized Neural Network Repair.
\newblock arXiv:2306.13793.

\bibitem[{Teuber et~al.(2025)Teuber, Kern, Janzen, and Beckert}]{Teuber2025}
Teuber, S.; Kern, P.; Janzen, M.; and Beckert, B. 2025.
\newblock Revisiting Differential Verification: Equivalence Verification with Confidence.
\newblock In \emph{Tools and Algorithms for the Construction and Analysis of Systems}, volume 15697 of \emph{Lecture Notes in Computer Science}, 257--278. Springer.

\bibitem[{Zhang, Song, and Sun(2023)}]{Zhang2023}
Zhang, Y.; Song, F.; and Sun, J. 2023.
\newblock {QEBVerif}: Quantization Error Bound Verification of Neural Networks.
\newblock In \emph{Computer Aided Verification}, volume 13965 of \emph{Lecture Notes in Computer Science}, 413--437. Springer.

\bibitem[{Zhang et~al.(2022)Zhang, Zhao, Chen, Song, Zhang, Chen, and Sun}]{Zhang2022}
Zhang, Y.; Zhao, Z.; Chen, G.; Song, F.; Zhang, M.; Chen, T.; and Sun, J. 2022.
\newblock {QVIP}: An {ILP}-based Formal Verification Approach for Quantized Neural Networks.
\newblock In \emph{Proceedings of the 37th IEEE/ACM International Conference on Automated Software Engineering}, 82:1--82:13. ACM.

\end{thebibliography}

\clearpage
\appendix
\setcounter{secnumdepth}{1}
\section{Abstract Round-Clamp: Complete Formulation}
\label{app:round_clamp}

\noindent Here, $\lfloor x\rceil$ denotes rounding to the nearest integer, with ties broken toward the even integer. We define $r_l = \lfloor l\rceil$ and $r_u = \min(\lfloor u\rceil,C)$.

\noindent\textbf{Degenerate hull edges.}
In Cases~3b, 5, and~6, define
\[
d_l=r_l+\tfrac12-l,\qquad
d_u=u+\tfrac12-r_u.
\]
The applicable oblique hull constraints can be
written without division as
\[
d_l(y-r_l)\leq x-l,\qquad
d_u(y-r_u)\geq x-u.
\]
For positive denominators, these are equivalent
to the slope formulations below. When a denominator
vanishes, the corresponding constraint reduces
to $x\geq l$ or $x\leq u$ and is redundant.
The slope expressions below are therefore used
only for positive denominators.

\subsubsection{Case 1: Fully Below Lower Threshold ($u \leq -1/2$)}

\noindent The entire input range rounds to values below zero, which are then clamped to zero.\\

\noindent \textbf{Parameters:}
\begin{equation*}
\widehat{\lambda} = 0, \quad \widehat{\mu} = 0, \quad \widehat{e} = 0
\end{equation*}

\noindent \textbf{Zonotope update:}
\begin{equation*}
\mathbf{v}_i^{out} = 0, \quad \mathbf{M}_{i,:}^{out} = \mathbf{0}
\end{equation*}

\medskip

\noindent \textbf{Constrained zonotopes:} No additional constraints needed.

\noindent \subsubsection{Case 2: Fully Above Upper Threshold ($l \geq C + 1/2$)}

The entire input range rounds to values above $C$, which are then clamped to $C$.

\noindent \textbf{Parameters:}
\begin{equation*}
\widehat{\lambda} = 0, \quad \widehat{\mu} = C, \quad \widehat{e} = 0
\end{equation*}

\noindent \textbf{Zonotope update:}
\begin{equation*}
\mathbf{v}_i^{out} = C, \quad \mathbf{M}_{i,:}^{out} = \mathbf{0}
\end{equation*}

\noindent \textbf{Constrained zonotopes:} No additional constraints needed.

\subsubsection{Case 3: Within Clamp Range
($-\tfrac12 \leq l < C+\tfrac12$,
$-\tfrac12 < u \leq C+\tfrac12$)}

The input interval lies within the effective clamp range, with upper clamping possible at
$u=C+\tfrac12$.

\noindent \textbf{Parameters:} We distinguish whether the interval crosses a rounding threshold.\\

\noindent If $r_l = r_u$, then the rounding function is constant on $[l,u]$:
\begin{equation*}
\widehat{\lambda} = 0, \quad \widehat{\mu} = r_l, \quad \widehat{e} = 0.
\end{equation*}

\noindent If $r_l < r_u$, then at least one rounding threshold is crossed:
\begin{equation*}
\widehat{\lambda} = 1, \quad \widehat{\mu} = 0, \quad \widehat{e} = 0.5.
\end{equation*}

\noindent \textbf{Zonotope update:} If $r_l = r_u$, then
\begin{equation*}
\mathbf{v}_i^{out} = r_l, \quad \mathbf{M}_{i,:}^{out} = \mathbf{0}.
\end{equation*}

\noindent If $r_l < r_u$, then
\begin{equation*}
\mathbf{v}_i^{out} = \mathbf{v}_i, \quad \mathbf{M}_{i,:}^{out} = (\mathbf{M}_{i,:}, \widehat{e}).
\end{equation*}

\noindent \textbf{Constrained zonotopes:}
If $r_l = r_u$, no additional constraints are needed.
Otherwise, add the four inequalities below, using $d_l$ and
$d_u$ defined above. When either coefficient vanishes, the
corresponding oblique constraint is redundant.

\noindent Constraints:
\begin{align*}
\mathbf{A}_{K+1,:}
    &= (\mathbf{M}_{i,:}, \widehat{e}), \quad
\mathbf{b}_{K+1}
    = \mathbf{v}_i - r_l \\
\mathbf{A}_{K+2,:}
    &= -(\mathbf{M}_{i,:}, \widehat{e}), \quad
\mathbf{b}_{K+2}
    = -\mathbf{v}_i + r_u \\
\mathbf{A}_{K+3,:}
    &= ((1-d_l)\mathbf{M}_{i,:},-d_l\widehat e),\\
\mathbf{b}_{K+3}
    &= (1-d_l)\mathbf{v}_i-l+d_l r_l,\\
\mathbf{A}_{K+4,:}
    &= ((d_u-1)\mathbf{M}_{i,:},d_u\widehat e),\\
\mathbf{b}_{K+4}
    &= (d_u-1)\mathbf{v}_i+u-d_u r_u.
\end{align*}

\subsubsection{Case 4: Full Range
($l \leq -\tfrac12$, $u \geq C+\tfrac12$,
$(l,u)\neq(-\tfrac12,C+\tfrac12)$)}

\noindent The input range spans both rounding and clamping thresholds. Both effects must be captured.\\

\noindent \textbf{Parameters:} Select $\widehat{\lambda}$ to minimize
the width of the enclosing affine strip, illustrated in
Figure~\ref{fig:round_clamp_case4}. The candidate slopes are:
\begin{equation*}
\widehat{\lambda}_1 = \frac{C}{u - 0.5}, \quad
\widehat{\lambda}_2 = \frac{C}{C - l - 0.5}.
\end{equation*}

\noindent Here $\widehat{\lambda}_1$ defines the lower hull edge
\[
L(x)=\widehat{\lambda}_1\left(x-\frac12\right),
\]
and $\widehat{\lambda}_2$ defines the upper hull edge
\[
U(x)=\widehat{\lambda}_2(x-l).
\]

\noindent Selection criterion: If $C - l \leq u$, use
$\widehat{\lambda} = \widehat{\lambda}_1$; otherwise use
$\widehat{\lambda} = \widehat{\lambda}_2$.\\

\noindent Error bounds:
\begin{align*}
\widehat{\mu}_1
    &= \max(-\widehat{\lambda} l,\,
        C - \widehat{\lambda}(C - 0.5)), \\
\widehat{\mu}_2
    &= \min(-0.5\widehat{\lambda},\,
        C - \widehat{\lambda} u).
\end{align*}

\noindent Final parameters:
\begin{equation*}
\widehat{\mu} = \frac{\widehat{\mu}_1 + \widehat{\mu}_2}{2}, \quad
\widehat{e} = \frac{\widehat{\mu}_1 - \widehat{\mu}_2}{2}.
\end{equation*}

\noindent \textbf{Zonotope update:}
\begin{equation*}
\mathbf{v}_i^{out} = \widehat{\lambda} \mathbf{v}_i + \widehat{\mu}, \quad
\mathbf{M}_{i,:}^{out} = (\widehat{\lambda} \mathbf{M}_{i,:}, \widehat{e}).
\end{equation*}

\noindent \textbf{Constrained zonotopes:} Add three constraints. The first two
constraints encode
\[
y\geq 0,
\qquad
y\leq C.
\]
The third constraint adds the oblique hull edge not already given by
the selected affine strip.\\

\noindent Constraints:
\begin{align*}
\mathbf{A}_{K+1,:}
    &= (\widehat{\lambda}\mathbf{M}_{i,:}, \widehat{e}), \\
\mathbf{b}_{K+1}
    &= \widehat{\mu}+\widehat{\lambda}\mathbf{v}_i, \\[0.5em]
\mathbf{A}_{K+2,:}
    &= -(\widehat{\lambda}\mathbf{M}_{i,:}, \widehat{e}), \\
\mathbf{b}_{K+2}
    &= C-(\widehat{\mu}+\widehat{\lambda}\mathbf{v}_i).
\end{align*}

\noindent Third constraint. If $C-l\leq u$, then
$\widehat{\lambda}=\widehat{\lambda}_1$, so the affine strip already
contains the lower edge $L(x)$. We add the upper edge $y\leq U(x)$:
\begin{align*}
\mathbf{A}_{K+3,:}
    &= ((\widehat{\lambda}_2-\widehat{\lambda})\mathbf{M}_{i,:},
        -\widehat{e}), \\
\mathbf{b}_{K+3}
    &= (\widehat{\lambda}_2-\widehat{\lambda})\mathbf{v}_i
       -\widehat{\mu}
       -\widehat{\lambda}_2 l.
\end{align*}

\noindent If $C-l>u$, then $\widehat{\lambda}=\widehat{\lambda}_2$, so the
affine strip already contains the upper edge $U(x)$. We add the lower
edge $y\geq L(x)$:
\begin{align*}
\mathbf{A}_{K+3,:}
    &= ((\widehat{\lambda}-\widehat{\lambda}_1)\mathbf{M}_{i,:},
        \widehat{e}), \\
\mathbf{b}_{K+3}
    &= (\widehat{\lambda}-\widehat{\lambda}_1)\mathbf{v}_i
       +\widehat{\mu}
       +\frac{\widehat{\lambda}_1}{2}.
\end{align*}

\subsubsection{Case 5: Crosses Zero
($l < -\tfrac12$, $-\tfrac12 < u < C+\tfrac12$)}

\noindent The range crosses the lower rounding threshold but stays within the upper clamp bound.

\noindent \textbf{Parameters:} Let $r_u = \lfloor u \rceil$.
\begin{align*}
\widehat{\lambda}
    &= \frac{r_u}{r_u - l - 0.5}, \\
\widehat{\mu}_1
    &= -\widehat{\lambda}l
     = \frac{-r_u l}{r_u - l - 0.5}, \\
\widehat{\mu}_2
    &= -\frac{\widehat{\lambda}}{2}
     = -\frac{r_u/2}{r_u - l - 0.5}.
\end{align*}

\noindent Final parameters:
\begin{equation*}
\widehat{\mu} = \frac{\widehat{\mu}_1 + \widehat{\mu}_2}{2}, \quad
\widehat{e} = \frac{\widehat{\mu}_1 - \widehat{\mu}_2}{2}.
\end{equation*}

\noindent \textbf{Zonotope update:}
\begin{equation*}
\mathbf{v}_i^{out} = \widehat{\lambda} \mathbf{v}_i + \widehat{\mu}, \quad
\mathbf{M}_{i,:}^{out} = (\widehat{\lambda} \mathbf{M}_{i,:}, \widehat{e}).
\end{equation*}

\noindent \textbf{Constrained zonotopes:} Add the four inequalities below, using $d_u$ defined above.
When $d_u=0$, the fourth constraint is redundant.

\noindent Constraints:
\begin{align*} \mathbf{A}_{K+1,:} &= (\widehat{\lambda} \mathbf{M}_{i,:}, \widehat{e}), \\ \mathbf{b}_{K+1} &= \widehat{\mu} + \widehat{\lambda} \mathbf{v}_i \\ \mathbf{A}_{K+2,:} &= -(\widehat{\lambda} \mathbf{M}_{i,:}, \widehat{e}), \\ \mathbf{b}_{K+2} &= r_u - (\widehat{\mu} + \widehat{\lambda} \mathbf{v}_i) \\ \mathbf{A}_{K+3,:} &= ((\widehat{\lambda}-1)\mathbf{M}_{i,:}, \widehat{e}), \\ \mathbf{b}_{K+3} &= (\widehat{\lambda}-1)\mathbf{v}_i + \widehat{\mu} + \frac12 \\ \mathbf{A}_{K+4,:}
    &= ((d_u\widehat{\lambda}-1)\mathbf{M}_{i,:},
        d_u\widehat e),\\
\mathbf{b}_{K+4}
    &= (d_u\widehat{\lambda}-1)\mathbf{v}_i
       +d_u(\widehat{\mu}-r_u)+u. \end{align*}

\subsubsection{Case 6: Crosses Upper Clamp
($-\tfrac12 < l < C+\tfrac12$,
$u > C+\tfrac12$)}

\noindent The range starts within clamp bounds but exceeds the upper clamp threshold.\\

\noindent \textbf{Parameters:} Let $r_l = \lfloor l \rceil$.
\begin{align*}
\widehat{\lambda}
    &= \frac{C - r_l}{u - r_l - 0.5}, \\
\widehat{\mu}_1
    &= C - \widehat{\lambda}(C - 0.5), \\
\widehat{\mu}_2
    &= C - \widehat{\lambda} u.
\end{align*}

\noindent Final parameters:
\begin{equation*}
\widehat{\mu} = \frac{\widehat{\mu}_1 + \widehat{\mu}_2}{2}, \quad
\widehat{e} = \frac{\widehat{\mu}_1 - \widehat{\mu}_2}{2}.
\end{equation*}

\noindent \textbf{Zonotope update:}
\begin{equation*}
\mathbf{v}_i^{out} = \widehat{\lambda} \mathbf{v}_i + \widehat{\mu}, \quad
\mathbf{M}_{i,:}^{out} = (\widehat{\lambda} \mathbf{M}_{i,:}, \widehat{e}).
\end{equation*}

\noindent \textbf{Constrained zonotopes:}
Add the four inequalities below, using $d_l$ defined above.
When $d_l=0$, the fourth constraint is redundant.

\noindent Constraints:
\begin{align*} \mathbf{A}_{K+1,:} &= (\widehat{\lambda} \mathbf{M}_{i,:}, \widehat{e}), \\ \mathbf{b}_{K+1} &= \widehat{\mu} + \widehat{\lambda} \mathbf{v}_i - r_l \\ \mathbf{A}_{K+2,:} &= -(\widehat{\lambda} \mathbf{M}_{i,:}, \widehat{e}), \\ \mathbf{b}_{K+2} &= C - (\widehat{\mu} + \widehat{\lambda} \mathbf{v}_i) \\ \mathbf{A}_{K+3,:} &= ((1-\widehat{\lambda})\mathbf{M}_{i,:}, -\widehat{e}), \\ \mathbf{b}_{K+3} &= (1-\widehat{\lambda})\mathbf{v}_i - \widehat{\mu} + \frac12 \\ \mathbf{A}_{K+4,:}
    &= ((1-d_l\widehat{\lambda})\mathbf{M}_{i,:},
        -d_l\widehat e),\\
\mathbf{b}_{K+4}
    &= (1-d_l\widehat{\lambda})\mathbf{v}_i
       -d_l(\widehat{\mu}-r_l)-l. \end{align*}

\section{Soundness and Optimality of the Round-Clamp
Abstract Transformer}
\label{app:soundness_optimality}

We first prove that the abstract transformer for
$f(x)=\mathrm{clamp}(\lfloor x\rceil,0,C)$ is \textit{sound} and \textit{optimal} at the
level of the unconstrained affine strip. More precisely, for each fixed
slope $\lambda$, the chosen offset $\widehat{\mu}$ and error radius
$\widehat e$ give the tightest possible vertical interval abstraction,
and the selected slope $\widehat{\lambda}$ minimizes this interval width.
This is an optimality statement for the affine strip only. The stronger
geometric optimality statement for the constrained-zonotope refinement,
namely convex-hull optimality of the induced polygon, is proved
separately in Appendix~\ref{app:constrained_soundness}.

\subsection*{Definitions}

For a fixed slope $\lambda \geq 0$, define the \emph{residual}
$g_\lambda : [l,u] \to \mathbb{R}$ by $g_\lambda(x) = f(x) - \lambda x$.
The tightest sound affine abstraction with slope $\lambda$ has:
\begin{align}
\mu^*(\lambda) &= \tfrac{1}{2}\bigl(
    \sup_{x \in [l,u]} g_\lambda(x) 
    + \inf_{x \in [l,u]} g_\lambda(x)\bigr), 
    \label{eq:opt_mu} \\
e^*(\lambda) &= \tfrac{1}{2}\bigl(
    \sup_{x \in [l,u]} g_\lambda(x) 
    - \inf_{x \in [l,u]} g_\lambda(x)\bigr).
    \label{eq:opt_e}
\end{align}

\noindent The \emph{optimal slope} minimizes the abstraction width:
\begin{equation}
\widehat{\lambda} \in \arg\min_{\lambda \geq 0}\ e^*(\lambda).
\label{eq:opt_lambda}
\end{equation}

\noindent An abstraction $(\widehat{\lambda}, \widehat{\mu}, \widehat{e})$ 
is a \emph{tight overapproximation} if it is sound and 
$\widehat{\mu} = \mu^*(\widehat{\lambda})$, 
$\widehat{e} = e^*(\widehat{\lambda})$, 
with $\widehat{\lambda}$ satisfying \eqref{eq:opt_lambda}.

\begin{lemma}[Soundness from Optimality]
\label{lem:soundness}
If $\widehat{\mu} = \mu^*(\widehat{\lambda})$ and 
$\widehat{e} = e^*(\widehat{\lambda})$ as defined in 
\eqref{eq:opt_mu}-\eqref{eq:opt_e}, then the abstraction 
is sound: for all $x \in [l,u]$,
\[
f(x) \in [\widehat{\lambda}x + \widehat{\mu} - \widehat{e},\ 
           \widehat{\lambda}x + \widehat{\mu} + \widehat{e}].
\]
\end{lemma}

\begin{proof}
By definition of $\mu^*$ and $e^*$:
\begin{align*}
\widehat{\lambda}x + \widehat{\mu} + \widehat{e} 
&= \widehat{\lambda}x + \sup_{t \in [l,u]} g_{\widehat{\lambda}}(t) \\
&\geq \widehat{\lambda}x + g_{\widehat{\lambda}}(x) \\
&= \widehat{\lambda}x + f(x) - \widehat{\lambda}x = f(x).
\end{align*}
\begin{align*}
\widehat{\lambda}x + \widehat{\mu} - \widehat{e} 
&= \widehat{\lambda}x + \inf_{t \in [l,u]} g_{\widehat{\lambda}}(t) \\
&\leq \widehat{\lambda}x + g_{\widehat{\lambda}}(x) = f(x).
\end{align*}
\end{proof}

\begin{theorem}[Soundness and Tight Overapproximation]
\label{thm:main}
For each case, the parameters $(\widehat{\lambda}, \widehat{\mu}, 
\widehat{e})$ defined in the round-clamp abstract transformer 
satisfy:
\begin{enumerate}
    \item \textbf{Optimality}: $\widehat{\lambda}$ minimizes 
    $e^*(\lambda)$ over all $\lambda \geq 0$, and 
    $\widehat{\mu} = \mu^*(\widehat{\lambda})$, 
    $\widehat{e} = e^*(\widehat{\lambda})$.
    \item \textbf{Soundness}: 
    for all $x \in [l,u]$,
    \[
    f(x) \in [\widehat{\lambda}x + \widehat{\mu} - \widehat{e},\ 
               \widehat{\lambda}x + \widehat{\mu} + \widehat{e}].
    \]
\end{enumerate}
\end{theorem}

\begin{proof}
We use throughout the rounding inequality:
\begin{equation}
x - \tfrac{1}{2} \leq \lfloor x \rceil \leq x + \tfrac{1}{2}
\quad \forall x \in \mathbb{R}.
\tag{R}\label{eq:round} 
\end{equation} 
For each case, we (1) compute $\sup g_\lambda$ and $\inf g_\lambda$ explicitly, (2) verify that $\widehat{\mu} = \mu^*(\widehat{\lambda})$ and $\widehat{e} = e^*(\widehat{\lambda})$ as defined in \eqref{eq:opt_mu}-\eqref{eq:opt_e}, and (3) show $\widehat{\lambda}$ minimizes $e^*(\lambda)$. Soundness then follows immediately from Lemma~\ref{lem:soundness}. Throughout the proof, we consider the one-sided limits
at rounding thresholds when computing extrema.
This ensures that half-integer boundaries are
handled correctly.

\paragraph{Cases 1-2 (trivial).}

$f$ is constant on $[l,u]$, so $g_\lambda(x) = C_0 - \lambda x$ for some constant $C_0 \in \{0, C\}$. The unique optimal slope is $\widehat{\lambda} = 0$, which gives $e^*(0) = 0$, $\mu^*(0) = C_0$. Any $\lambda > 0$ gives $e^*(\lambda) = \frac{\lambda(u-l)}{2} > 0$. Our parameters $\widehat{\lambda}=0$, $\widehat{\mu}=C_0$, 
$\widehat{e}=0$ are optimal. \checkmark

\paragraph{Case 3: $-\tfrac12 \leq l < C+\tfrac12$, $-\tfrac12 < u \leq C+\tfrac12$.}

In Case 3, the round-clamp function satisfies
\[
f(x)=\min(\lfloor x\rceil,C),
\qquad
g_\lambda(x)=f(x)-\lambda x.
\]
Let
\[
r_l=\lfloor l\rceil,
\qquad
r_u=\min(\lfloor u\rceil,C).
\]
We distinguish two subcases.

\medskip
\noindent
\textbf{Case 3a: $r_l=r_u$.}\\

\noindent In this case no rounding threshold is crossed, so $f$ is constant on
$[l,u]$:
\[
f(x)=r_l
\qquad \forall x\in[l,u].
\]
Thus, for any $\lambda\geq 0$,
\[
g_\lambda(x)=r_l-\lambda x.
\]
Since $g_\lambda$ is decreasing on $[l,u]$, we have
\[
\sup_{x\in[l,u]} g_\lambda(x)=r_l-\lambda l,
\qquad
\inf_{x\in[l,u]} g_\lambda(x)=r_l-\lambda u.
\]
Therefore
\[
\mu^*(\lambda)
=
\frac12\left(
r_l-\lambda l+r_l-\lambda u
\right)
=
r_l-\frac{\lambda(l+u)}{2},
\]
and
\[
e^*(\lambda)
=
\frac12\left(
r_l-\lambda l-(r_l-\lambda u)
\right)
=
\frac{\lambda(u-l)}{2}.
\]
The function $e^*(\lambda)$ is minimized over $\lambda\geq 0$ at
\[
\widehat{\lambda}=0.
\]
For this choice,
\[
\mu^*(0)=r_l,
\qquad
e^*(0)=0.
\]
Hence, in Case 3a, the optimal parameters are
\[
\widehat{\lambda}=0,
\qquad
\widehat{\mu}=r_l,
\qquad
\widehat e=0.
\]\checkmark

\medskip
\noindent
\textbf{Case 3b: $r_l<r_u$.}\\

\noindent In this case at least one rounding threshold is crossed. For
$k\in\{r_l,\dots,r_u\}$, define the portion of the $k$-th rounding bin
intersected by $[l,u]$ as
\[
I_k
=
[l,u]\cap [k-\tfrac12,k+\tfrac12].
\]
Equivalently, write
\[
a_k=\max\{l,k-\tfrac12\},
\qquad
b_k=\min\{u,k+\tfrac12\},
\]
so that $I_k=[a_k,b_k]$ whenever this intersection is nonempty. On the interior of each rounding bin, the function is constant and equal to $k$. The extrema below also account for one-sided limits at bin boundaries. Thus,
\[
g_\lambda(x)=k-\lambda x.
\]
Since $\lambda\geq0$, this residual is decreasing on each bin. Hence
\[
\sup_{x\in I_k} g_\lambda(x)=k-\lambda a_k,
\qquad
\inf_{x\in I_k} g_\lambda(x)=k-\lambda b_k.
\]
Taking the maximum and minimum over all active bins gives
\[
\sup_{x\in[l,u]} g_\lambda(x)
=
\max_{k=r_l,\dots,r_u}
\left(k-\lambda a_k\right),
\]
and
\[
\inf_{x\in[l,u]} g_\lambda(x)
=
\min_{k=r_l,\dots,r_u}
\left(k-\lambda b_k\right).
\]

\noindent We now evaluate these quantities at $\lambda=1$. For every active bin,
\[
a_k\geq k-\frac12
\qquad\text{and}\qquad
b_k\leq k+\frac12.
\]
Therefore
\[
k-a_k\leq \frac12,
\qquad
k-b_k\geq -\frac12.
\]
Since $r_l<r_u$, the interval $[l,u]$ crosses at least one rounding
threshold. Hence there exists an integer $k$ such that
\[
k+\frac12\in[l,u].
\]
At this threshold, the adjacent residual values for $\lambda=1$ attain
the limiting values
\[
k-\left(k+\frac12\right)=-\frac12,
\qquad
(k+1)-\left(k+\frac12\right)=\frac12.
\]
Thus
\[
\sup_{x\in[l,u]} g_1(x)=\frac12,
\qquad
\inf_{x\in[l,u]} g_1(x)=-\frac12.
\]
Consequently,
\[
\mu^*(1)
=
\frac12\left(\frac12-\frac12\right)
=
0,
\]
and
\[
e^*(1)
=
\frac12\left(\frac12-(-\frac12)\right)
=
\frac12.
\]

\noindent It remains to show that $\lambda=1$ minimizes $e^*(\lambda)$. Let
$\lambda\geq0$ be arbitrary. Since $r_l<r_u$, there exists a rounding
threshold $h=k+\frac12$ crossed by $[l,u]$. Across this threshold, the
rounding function jumps from $k$ to $k+1$. Therefore the residual
$g_\lambda$ has adjacent limiting values
\[
k-\lambda h
\qquad\text{and}\qquad
k+1-\lambda h.
\]
These two values differ by exactly $1$. Hence
\[
\sup_{x\in[l,u]} g_\lambda(x)
-
\inf_{x\in[l,u]} g_\lambda(x)
\geq 1.
\]
Therefore
\[
e^*(\lambda)
=
\frac12\left(
\sup_{x\in[l,u]} g_\lambda(x)
-
\inf_{x\in[l,u]} g_\lambda(x)
\right)
\geq \frac12.
\]
Since $e^*(1)=1/2$, this proves that
\[
\widehat{\lambda}=1
\]
minimizes $e^*(\lambda)$ over $\lambda\geq0$. Hence, in Case 3b, the
optimal parameters are
\[
\widehat{\lambda}=1,
\qquad
\widehat{\mu}=0,
\qquad
\widehat e=\frac12.
\]
\checkmark

\medskip
\noindent
Combining the two subcases, the optimal Case 3 parameters are
\[
(\widehat{\lambda},\widehat{\mu},\widehat e)
=
\begin{cases}
(0,r_l,0), & r_l=r_u,\\[1mm]
(1,0,\frac12), & r_l<r_u.
\end{cases}
\]

\paragraph{Case 4: $l \leq -\tfrac12$,
$u \geq C+\tfrac12$,
$(l,u)\neq(-\tfrac12,C+\tfrac12)$.}

The input interval spans the lower clamp region, the rounding region,
and the upper clamp region. We assume $C>0$; the case $C=0$ is
constant and reduces to Cases 1--2. Define
\[
g_\lambda(x)=f(x)-\lambda x.
\]
We prove optimality by computing the residual extrema, verifying
$\widehat{\mu}=\mu^*(\widehat{\lambda})$ and
$\widehat e=e^*(\widehat{\lambda})$, and then minimizing
$e^*(\lambda)$ over $\lambda\geq0$.

\medskip
\noindent
\textbf{Step 1: Compute $\sup g_\lambda$ and $\inf g_\lambda$.}\\

\noindent First consider $0\leq \lambda\leq 1$. On the lower clamped region,
$f(x)=0$, so the largest residual is obtained at the left endpoint:
\[
\sup_{x\leq -1/2} g_\lambda(x)=-\lambda l.
\]
On the upper clamped region, $f(x)=C$, so the smallest residual is
obtained at the right endpoint:
\[
\inf_{x\geq C+1/2} g_\lambda(x)=C-\lambda u.
\]

\noindent Inside the unclamped rounding region, let
\[
R_2=[-1/2,C+1/2].
\]
For each integer $k\in\{0,\dots,C\}$, define the $k$-th rounding
plateau
\[
I_k=[k-\tfrac12,k+\tfrac12],
\]
with the obvious truncation at the endpoints of $R_2$. On $I_k$,
\[
f(x)=k,
\qquad
g_\lambda(x)=k-\lambda x.
\]
Since $\lambda\geq0$, the function $g_\lambda$ is decreasing on each
plateau. Therefore, on the plateau $I_k$, the largest residual is
attained at the left endpoint and the smallest residual is attained at
the right endpoint:
\[
\sup_{x\in I_k} g_\lambda(x)
=
k-\lambda\left(k-\frac12\right),
\]
and
\[
\inf_{x\in I_k} g_\lambda(x)
=
k-\lambda\left(k+\frac12\right).
\]
Equivalently,
\[
\sup_{x\in I_k} g_\lambda(x)
=
(1-\lambda)k+\frac{\lambda}{2},
\]
and
\[
\inf_{x\in I_k} g_\lambda(x)
=
(1-\lambda)k-\frac{\lambda}{2}.
\]

\noindent Because $0\leq\lambda\leq1$, both expressions are nondecreasing in
$k$. Hence, over the whole rounding region $R_2$, the largest residual
is attained on the last plateau, $k=C$, while the smallest residual is
attained on the first plateau, $k=0$. Thus
\[
\sup_{x\in R_2} g_\lambda(x)
=
C-\lambda\left(C-\frac12\right),
\]
and
\[
\inf_{x\in R_2} g_\lambda(x)
=
-\frac{\lambda}{2}.
\]

\noindent On the lower clamped region $R_1=[l,-1/2]$, we have $f(x)=0$, so
\[
g_\lambda(x)=-\lambda x.
\]
Thus
\[
\sup_{x\in R_1} g_\lambda(x)=-\lambda l,
\qquad
\inf_{x\in R_1} g_\lambda(x)=\frac{\lambda}{2}.
\]

\noindent On the upper clamped region $R_3=[C+1/2,u]$, we have $f(x)=C$, so
\[
g_\lambda(x)=C-\lambda x.
\]
Thus
\[
\sup_{x\in R_3} g_\lambda(x)
=
C-\lambda\left(C+\frac12\right),
\qquad
\inf_{x\in R_3} g_\lambda(x)
=
C-\lambda u.
\]

\noindent Combining the three regions gives
\[
\begin{aligned}
\sup_{x\in[l,u]} g_\lambda(x)
=
\max\Bigl\{&
-\lambda l,\\
& C-\lambda\Bigl(C-\frac12\Bigr),\\
& C-\lambda\Bigl(C+\frac12\Bigr)
\Bigr\}.
\end{aligned}
\]
Since
\[
C-\lambda\left(C-\frac12\right)
\geq
C-\lambda\left(C+\frac12\right),
\]
the third term is dominated. Hence
\[
\sup_{x\in[l,u]} g_\lambda(x)
=
\max\left\{
-\lambda l,\,
C-\lambda\left(C-\frac12\right)
\right\}.
\]

\noindent Similarly,
\[
\inf_{x\in[l,u]} g_\lambda(x)
=
\min\left\{
\frac{\lambda}{2},\,
-\frac{\lambda}{2},\,
C-\lambda u
\right\}.
\]
Since $\lambda\geq0$, we have $-\lambda/2\leq \lambda/2$, so
\[
\inf_{x\in[l,u]} g_\lambda(x)
=
\min\left\{
-\frac{\lambda}{2},\,
C-\lambda u
\right\}.
\]

\noindent Indeed, the remaining candidates are dominated in this range of
$\lambda$: since $l\leq -1/2$,
\[
-\lambda l\geq \frac{\lambda}{2},
\]
and since $\lambda\leq1$,
\[
C-\lambda\left(C-\frac12\right)
\geq
-\frac{\lambda}{2}.
\]

\noindent Therefore, for $0\leq\lambda\leq1$,
\begin{align*}
2e^*(\lambda)
&=
\max\Bigl\{
-\lambda l,\,
C-\lambda\Bigl(C-\frac12\Bigr)
\Bigr\} \\
&\quad
-
\min\Bigl\{
-\frac{\lambda}{2},\,
C-\lambda u
\Bigr\}.
\end{align*}

\noindent It is useful to rewrite this as
\[
2e^*(\lambda)
=
S(\lambda)+T(\lambda),
\]
where
\[
S(\lambda)
=
\max\left\{
-\lambda l,\,
C-\lambda\left(C-\frac12\right)
\right\},
\]
and
\[
T(\lambda)
=
\max\left\{
\frac{\lambda}{2},\,
\lambda u-C
\right\}.
\]

\medskip
\noindent
\textbf{Step 2: Verify $\widehat{\mu}$ and $\widehat e$.}\\

\noindent The two breakpoints of $S$ and $T$ are
\[
\widehat{\lambda}_2
=
\frac{C}{C-l-\frac12},
\qquad
\widehat{\lambda}_1
=
\frac{C}{u-\frac12}.
\]
Since $l\leq -1/2$ and $u\geq C+1/2$, both satisfy
\[
0\leq \widehat{\lambda}_1,\widehat{\lambda}_2\leq1.
\]

\noindent Let $\widehat{\lambda}=\min\{\widehat{\lambda}_1,
\widehat{\lambda}_2\}$. Substituting $\widehat{\lambda}$ into the residual extrema gives
\[
\sup_{x\in[l,u]} g_{\widehat{\lambda}}(x)
=
\max\left\{
-\widehat{\lambda}l,\,
C-\widehat{\lambda}\left(C-\frac12\right)
\right\},
\]
and
\[
\inf_{x\in[l,u]} g_{\widehat{\lambda}}(x)
=
\min\left\{
-\frac{\widehat{\lambda}}{2},\,
C-\widehat{\lambda}u
\right\}.
\]
Thus, with the appendix notation,
\[
\widehat{\mu}_1
=
\sup_{x\in[l,u]} g_{\widehat{\lambda}}(x),
\qquad
\widehat{\mu}_2
=
\inf_{x\in[l,u]} g_{\widehat{\lambda}}(x),
\]
Therefore
\[
\widehat{\mu}
=
\frac{\widehat{\mu}_1+\widehat{\mu}_2}{2}
=
\mu^*(\widehat{\lambda}),
\]
and
\[
\widehat e
=
\frac{\widehat{\mu}_1-\widehat{\mu}_2}{2}
=
e^*(\widehat{\lambda}).
\]

\medskip
\noindent
\textbf{Step 3: Show that $\widehat{\lambda}$ minimizes $e^*(\lambda)$.}\\

\noindent For $0\leq\lambda\leq1$, the function
\[
2e^*(\lambda)=S(\lambda)+T(\lambda)
\]
is convex and piecewise linear. The breakpoint of $S$ is
\[
\widehat{\lambda}_2
=
\frac{C}{C-l-\frac12},
\]
and the breakpoint of $T$ is
\[
\widehat{\lambda}_1
=
\frac{C}{u-\frac12}.
\]

\noindent If $\widehat{\lambda}_1\leq \widehat{\lambda}_2$, equivalently
$C-l\leq u$, then for
$\lambda\in(\widehat{\lambda}_1,\widehat{\lambda}_2)$ we have
\[
S(\lambda)=C-\lambda\left(C-\frac12\right),
\qquad
T(\lambda)=\lambda u-C.
\]
Hence
\[
\frac{d}{d\lambda}
\bigl(S(\lambda)+T(\lambda)\bigr)
=
u-C+\frac12
\geq 1.
\]
Before $\widehat{\lambda}_1$, the function is nonincreasing, and after
$\widehat{\lambda}_1$ it is increasing. Thus a minimizer is
\[
\widehat{\lambda}=\widehat{\lambda}_1.
\]

\noindent If $\widehat{\lambda}_2<\widehat{\lambda}_1$, equivalently
$C-l>u$, then for
$\lambda\in(\widehat{\lambda}_2,\widehat{\lambda}_1)$ we have
\[
S(\lambda)=-\lambda l,
\qquad
T(\lambda)=\frac{\lambda}{2}.
\]
Hence
\[
\frac{d}{d\lambda}
\bigl(S(\lambda)+T(\lambda)\bigr)
=
-l+\frac12
\geq 1.
\]
Before $\widehat{\lambda}_2$, the function is nonincreasing, and after
$\widehat{\lambda}_2$ it is increasing. Thus a minimizer is
\[
\widehat{\lambda}=\widehat{\lambda}_2.
\]

\noindent It remains only to exclude $\lambda>1$. For $\lambda\geq1$, the maximum
residual is attained at the left endpoint and the minimum residual is
attained at the right endpoint:
\[
\sup_{x\in[l,u]}g_\lambda(x)=-\lambda l,
\qquad
\inf_{x\in[l,u]}g_\lambda(x)=C-\lambda u.
\]
Therefore
\[
2e^*(\lambda)
=
-\lambda l-\bigl(C-\lambda u\bigr)
=
\lambda(u-l)-C,
\]
which is increasing in $\lambda$. Hence no $\lambda>1$ improves on the
minimum attained in $[0,1]$.\\

\noindent Consequently,
\[
\widehat{\lambda}
=
\begin{cases}
\dfrac{C}{u-\frac12}, & C-l\leq u,\\[3mm]
\dfrac{C}{C-l-\frac12}, & C-l>u
\end{cases}
\]
minimizes $e^*(\lambda)$ over all $\lambda\geq0$. Together with Step 2,
this proves the optimality of the Case 4 parameters. \checkmark

\paragraph{Case 5: $l < -\tfrac12$,
$-\tfrac12 < u < C+\tfrac12$.}

The input interval crosses the lower clamp threshold but does not cross
the upper clamp threshold. Let
\[
r_u=\lfloor u\rceil.
\]
The function is clamped to $0$ on the left and then follows the
rounding staircase up to the value $r_u$.

\noindent If $r_u=0$, then $f$ is constant equal to $0$ on $[l,u]$, and the claim reduces to Cases 1-2. Assume therefore that $r_u\geq1$.

\medskip
\noindent
\textbf{Step 1: Compute $\sup g_\lambda$ and $\inf g_\lambda$.}\\

\noindent First consider $0\leq\lambda\leq1$. On the lower clamped region,
$f(x)=0$, so
\[
g_\lambda(x)=-\lambda x.
\]
Therefore the largest residual on the lower clamped region is
\[
-\lambda l.
\]

\noindent  Inside the unclamped rounding region, for each integer
$k\in\{0,\dots,r_u\}$, the $k$-th rounding plateau satisfies
\[
f(x)=k,
\qquad
g_\lambda(x)=k-\lambda x.
\]
Since $\lambda\geq0$, the residual is decreasing on each plateau.
Thus the largest residual on the $k$-th plateau is attained at its left
endpoint:
\[
k-\lambda\left(k-\frac12\right).
\]
For $0\leq\lambda\leq1$, this expression is nondecreasing in $k$.
Hence the largest residual in the rounding region is attained on the
last plateau, $k=r_u$:
\[
r_u-\lambda\left(r_u-\frac12\right).
\]

\noindent Thus
\[
S(\lambda)
=\sup_{x\in[l,u]} g_\lambda(x)
=
\max\left\{
-\lambda l,\,
r_u-\lambda\left(r_u-\frac12\right)
\right\}.
\]

\noindent For the infimum, the lower clamped region gives only residuals at least $\lambda/2$ near the boundary $x=-1/2$. The smallest residual in the
rounding region comes from the first plateau, $k=0$, whose right endpoint
gives
\[
-\frac{\lambda}{2}.
\]
The endpoint $u$ gives the additional candidate
\[
r_u-\lambda u.
\]
Therefore
\[
I(\lambda) = \inf_{x\in[l,u]} g_\lambda(x)
=
\min\left\{
-\frac{\lambda}{2},\,
r_u-\lambda u
\right\}.
\]

\noindent Thus, for $0\leq\lambda\leq1$,
\[
2e^*(\lambda)=S(\lambda)-I(\lambda),
\]

\medskip
\noindent
\textbf{Step 2: Verify $\widehat{\mu}$ and $\widehat e$.}\\

\noindent The breakpoint of $S$ is obtained by solving
\[
-\lambda l
=
r_u-\lambda\left(r_u-\frac12\right).
\]
This gives
\[
\widehat{\lambda}
=
\frac{r_u}{r_u-l-\frac12}.
\]
For this value of $\widehat{\lambda}$,
\[
-\widehat{\lambda}l
=
r_u-\widehat{\lambda}\left(r_u-\frac12\right),
\]
so
\[
\sup_{x\in[l,u]} g_{\widehat{\lambda}}(x)
=
-\widehat{\lambda}l.
\]
Moreover, since $u\leq r_u+\frac12$, we have
\[
\widehat{\lambda}\left(u-\frac12\right)
\leq r_u,
\]
and therefore
\[
-\frac{\widehat{\lambda}}{2}
\leq
r_u-\widehat{\lambda}u.
\]
Hence
\[
\inf_{x\in[l,u]} g_{\widehat{\lambda}}(x)
=
-\frac{\widehat{\lambda}}{2}.
\]

\noindent Thus, with the appendix notation,
\[
\widehat{\mu}_1
=
\sup_{x\in[l,u]}g_{\widehat{\lambda}}(x)
=
-\widehat{\lambda}l,
\]
and
\[
\widehat{\mu}_2
=
\inf_{x\in[l,u]}g_{\widehat{\lambda}}(x)
=
-\frac{\widehat{\lambda}}{2}.
\]
Equivalently,
\[
\widehat{\mu}_1
=
\frac{-r_u l}{r_u-l-\frac12},
\qquad
\widehat{\mu}_2
=
-\frac{r_u/2}{r_u-l-\frac12}.
\]
Therefore
\[
\widehat{\mu}
=
\frac{\widehat{\mu}_1+\widehat{\mu}_2}{2}
=
\mu^*(\widehat{\lambda}),
\]
and
\[
\widehat e
=
\frac{\widehat{\mu}_1-\widehat{\mu}_2}{2}
=
e^*(\widehat{\lambda}).
\]

\medskip
\noindent
\textbf{Step 3: Show that $\widehat{\lambda}$ minimizes $e^*(\lambda)$.}\\

\noindent For $0\leq\lambda\leq1$, we have
\[
2e^*(\lambda)=S(\lambda)-I(\lambda).
\]
The infimum is
\[
I(\lambda)
=
\min\left\{
-\frac{\lambda}{2},\,
r_u-\lambda u
\right\}.
\]
Since $u\leq r_u+\frac12$, for every
$0\leq\lambda\leq1$ we have
\[
r_u-\lambda u+\frac{\lambda}{2}
=
r_u-\lambda\left(u-\frac12\right)
\geq r_u(1-\lambda)
\geq 0.
\]
Consequently, $I(\lambda)=-\lambda/2$ throughout
$[0,1]$, and therefore
\[
2e^*(\lambda)
=
S(\lambda)+\frac{\lambda}{2}.
\]

\noindent The function $S$ is the maximum of two affine functions:
\[
S(\lambda)
=
\max\left\{
-\lambda l,\,
r_u-\lambda\left(r_u-\frac12\right)
\right\}.
\]
Its breakpoint is
\[
\widehat{\lambda}
=
\frac{r_u}{r_u-l-\frac12}.
\]
For $\lambda<\widehat{\lambda}$, the active term is
\[
r_u-\lambda\left(r_u-\frac12\right),
\]
so
\[
2e^*(\lambda)
=
r_u-\lambda\left(r_u-\frac12\right)
+
\frac{\lambda}{2}
=
r_u-\lambda(r_u-1),
\]
which is nonincreasing in $\lambda$.\\

\noindent For $\lambda>\widehat{\lambda}$, the active term is
\[
-\lambda l,
\]
so
\[
2e^*(\lambda)
=
-\lambda l+\frac{\lambda}{2}
=
\lambda\left(-l+\frac12\right),
\]
which is increasing because $l\leq -1/2$.\\

\noindent Thus the minimum is attained at the breakpoint
\[
\widehat{\lambda}
=
\frac{r_u}{r_u-l-\frac12}.
\]

\noindent It remains to exclude $\lambda>1$. Assume $r_u\geq1$ and let
$\lambda>1$. Since $l\leq -1/2$, the lower clamped region gives
\[
\sup_{x\in[l,u]} g_\lambda(x)
\geq
g_\lambda(l)
=
-\lambda l.
\]
Moreover, immediately before the last rounding jump, at
$x=r_u-\frac12$, the left limiting value of the residual is
\[
r_u-1-\lambda\left(r_u-\frac12\right).
\]
Hence
\[
\inf_{x\in[l,u]} g_\lambda(x)
\leq
r_u-1-\lambda\left(r_u-\frac12\right).
\]
Therefore
\[
2e^*(\lambda)
\geq
-\lambda l
-
\left(
r_u-1-\lambda\left(r_u-\frac12\right)
\right).
\]
Equivalently,
\[
2e^*(\lambda)
\geq
1-r_u+\lambda\left(r_u-l-\frac12\right).
\]
Since $\lambda>1$,
\[
2e^*(\lambda)
>
1-r_u+r_u-l-\frac12
=
\frac12-l.
\]
At the candidate slope,
\[
\widehat{\lambda}
=
\frac{r_u}{r_u-l-\frac12},
\]
we have
\[
2e^*(\widehat{\lambda})
=
\widehat{\lambda}\left(-l+\frac12\right).
\]
Because $l\leq -1/2$, we have
\[
\widehat{\lambda}
=
\frac{r_u}{r_u-l-\frac12}
\leq1.
\]
Thus
\[
2e^*(\widehat{\lambda})
\leq
-l+\frac12.
\]
Therefore,
\[
2e^*(\lambda)
>
2e^*(\widehat{\lambda})
\qquad
\forall \lambda>1.
\]
Hence no $\lambda>1$ improves on the candidate slope.\\

\noindent Consequently, the optimal Case 5 parameters are
\[
\widehat{\lambda}
=
\frac{r_u}{r_u-l-\frac12},
\]
\[
\widehat{\mu}_1
=
-\widehat{\lambda}l,
\qquad
\widehat{\mu}_2
=
-\frac{\widehat{\lambda}}{2},
\]
and
\[
\widehat{\mu}
=
\frac{\widehat{\mu}_1+\widehat{\mu}_2}{2},
\qquad
\widehat e
=
\frac{\widehat{\mu}_1-\widehat{\mu}_2}{2}.
\]
\checkmark

\paragraph{Case 6: $-\tfrac12 < l < C+\tfrac12$, $u > C+\tfrac12$.}

The input interval starts in the unclamped rounding region and crosses
the upper clamp threshold. Let
\[
r_l=\lfloor l\rceil.
\]

\noindent If $r_l=C$, then $f$ is constant equal to $C$ on $[l,u]$, and the claim reduces to Cases 1-2. Assume therefore that $r_l<C$.

\medskip
\noindent
\textbf{Step 1: Compute $\sup g_\lambda$ and $\inf g_\lambda$.}\\

\noindent First consider $0\leq\lambda\leq1$. Inside the unclamped rounding
region, for each integer $k\in\{r_l,\dots,C\}$, the $k$-th rounding
plateau satisfies
\[
f(x)=k,
\qquad
g_\lambda(x)=k-\lambda x.
\]
Since $\lambda\geq0$, the residual is decreasing on each plateau. Thus
the largest residual on the $k$-th plateau is attained at its left
endpoint:
\[
k-\lambda\left(k-\frac12\right),
\]
and the smallest residual is attained at its right endpoint:
\[
k-\lambda\left(k+\frac12\right).
\]
Equivalently,
\[
k-\lambda\left(k-\frac12\right)
=
(1-\lambda)k+\frac{\lambda}{2},
\]
and
\[
k-\lambda\left(k+\frac12\right)
=
(1-\lambda)k-\frac{\lambda}{2}.
\]
Because $0\leq\lambda\leq1$, both expressions are nondecreasing in
$k$. Therefore, in the rounding region, the largest residual is attained
on the last plateau, $k=C$, and the smallest residual is attained on the
first plateau, $k=r_l$.\\

\noindent Hence the rounding region contributes
\[
C-\lambda\left(C-\frac12\right)
\]
to the supremum, and
\[
r_l-\lambda\left(r_l+\frac12\right)
\]
to the infimum.\\

\noindent On the upper clamped region, $f(x)=C$, so
\[
g_\lambda(x)=C-\lambda x.
\]
Thus
\[
\sup_{x\geq C+1/2} g_\lambda(x)
=
C-\lambda\left(C+\frac12\right),
\]
and
\[
\inf_{x\geq C+1/2} g_\lambda(x)
=
C-\lambda u.
\]
The upper-clamp supremum is dominated by the rounding-region supremum,
since
\[
C-\lambda\left(C-\frac12\right)
\geq
C-\lambda\left(C+\frac12\right).
\]
Therefore,
\[
S(\lambda)
=
\sup_{x\in[l,u]} g_\lambda(x)
=
C-\lambda\left(C-\frac12\right),
\]
and
\[
I(\lambda)
=
\inf_{x\in[l,u]} g_\lambda(x)
=
\min\left\{
r_l-\lambda\left(r_l+\frac12\right),\,
C-\lambda u
\right\}.
\]

\noindent Thus, for $0\leq\lambda\leq1$,
\[
2e^*(\lambda)=S(\lambda)-I(\lambda),
\]
\medskip
\noindent
\textbf{Step 2: Verify $\widehat{\mu}$ and $\widehat e$.}\\

\noindent The breakpoint of $I$ is obtained by solving
\[
r_l-\lambda\left(r_l+\frac12\right)
=
C-\lambda u.
\]
This gives
\[
\widehat{\lambda}
=
\frac{C-r_l}{u-r_l-\frac12}.
\]
Since $u\geq C+1/2$, we have
\[
0\leq \widehat{\lambda}\leq1.
\]

\noindent For this value of $\widehat{\lambda}$,
\[
r_l-\widehat{\lambda}\left(r_l+\frac12\right)
=
C-\widehat{\lambda}u,
\]
so
\[
\inf_{x\in[l,u]}g_{\widehat{\lambda}}(x)
=
C-\widehat{\lambda}u.
\]
The supremum is
\[
\sup_{x\in[l,u]}g_{\widehat{\lambda}}(x)
=
C-\widehat{\lambda}\left(C-\frac12\right).
\]
Thus, with the appendix notation,
\[
\widehat{\mu}_1
=
\sup_{x\in[l,u]}g_{\widehat{\lambda}}(x)
=
C-\widehat{\lambda}\left(C-\frac12\right),
\]
and
\[
\widehat{\mu}_2
=
\inf_{x\in[l,u]}g_{\widehat{\lambda}}(x)
=
C-\widehat{\lambda}u.
\]
Equivalently,
\[
\widehat{\mu}_1
=
C-\frac{(C-r_l)(C-\frac12)}
        {u-r_l-\frac12},
\]
and
\[
\widehat{\mu}_2
=
C-\frac{(C-r_l)u}
        {u-r_l-\frac12}.
\]
Therefore
\[
\widehat{\mu}
=
\frac{\widehat{\mu}_1+\widehat{\mu}_2}{2}
=
\mu^*(\widehat{\lambda}),
\]
and
\[
\widehat e
=
\frac{\widehat{\mu}_1-\widehat{\mu}_2}{2}
=
e^*(\widehat{\lambda}).
\]

\medskip
\noindent
\textbf{Step 3: Show that $\widehat{\lambda}$ minimizes $e^*(\lambda)$.}\\

\noindent For $0\leq\lambda\leq1$, we have
\[
2e^*(\lambda)=S(\lambda)-I(\lambda).
\]
The supremum term is affine:
\[
S(\lambda)
=
C-\lambda\left(C-\frac12\right).
\]
The infimum term is the minimum of two affine functions:
\[
I(\lambda)
=
\min\left\{
r_l-\lambda\left(r_l+\frac12\right),\,
C-\lambda u
\right\}.
\]
Its breakpoint is
\[
\widehat{\lambda}
=
\frac{C-r_l}{u-r_l-\frac12}.
\]

\noindent For $\lambda<\widehat{\lambda}$, the active infimum term is
\[
r_l-\lambda\left(r_l+\frac12\right),
\]
so
\[
2e^*(\lambda)
=
C-\lambda\left(C-\frac12\right)
-
\left(
r_l-\lambda\left(r_l+\frac12\right)
\right).
\]
Equivalently,
\[
2e^*(\lambda)
=
C-r_l-\lambda(C-r_l-1).
\]
This is nonincreasing in $\lambda$.\\

\noindent For $\lambda>\widehat{\lambda}$, the active infimum term is
\[
C-\lambda u.
\]
Therefore
\[
2e^*(\lambda)
=
C-\lambda\left(C-\frac12\right)
-
(C-\lambda u).
\]
Equivalently,
\[
2e^*(\lambda)
=
\lambda\left(u-C+\frac12\right),
\]
which is increasing because $u\geq C+1/2$.\\

\noindent Thus the minimum over $0\leq\lambda\leq1$ is attained at the breakpoint
\[
\widehat{\lambda}
=
\frac{C-r_l}{u-r_l-\frac12}.
\]

\noindent It remains to exclude $\lambda>1$. Assume $r_l<C$ and let
$\lambda>1$. On the upper clamped region $f(x)=C$, so at the right
endpoint $x=u$,
\[
\inf_{x\in[l,u]} g_\lambda(x)
\leq
g_\lambda(u)
=
C-\lambda u.
\]
Immediately after the first rounding jump, at $x=r_l+\tfrac12$, the
right limiting value of the residual is
\[
(r_l+1)-\lambda\left(r_l+\tfrac12\right),
\]
so
\[
\sup_{x\in[l,u]} g_\lambda(x)
\geq
(r_l+1)-\lambda\left(r_l+\tfrac12\right).
\]
Therefore
\[
2e^*(\lambda)
\geq
\left[(r_l+1)-\lambda\left(r_l+\tfrac12\right)\right]
-
\left[C-\lambda u\right].
\]
Equivalently,
\[
2e^*(\lambda)
\geq
(r_l+1-C)+\lambda\left(u-r_l-\tfrac12\right).
\]
Since $\lambda>1$,
\[
2e^*(\lambda)
>
(r_l+1-C)+\left(u-r_l-\tfrac12\right)
=
u-C+\tfrac12.
\]
At the candidate slope
$\widehat{\lambda}=\dfrac{C-r_l}{u-r_l-\frac12}\leq1$, we have
\[
2e^*(\widehat{\lambda})
=
\widehat{\lambda}\left(u-C+\tfrac12\right)
\leq
u-C+\tfrac12.
\]
Therefore $2e^*(\lambda)>2e^*(\widehat{\lambda})$ for all $\lambda>1$,
so no $\lambda>1$ improves on the candidate slope.\\

\noindent Consequently, the optimal Case 6 parameters are
\[
\widehat{\lambda}
=
\frac{C-r_l}{u-r_l-\frac12},
\]
\[
\widehat{\mu}_1
=
C-\widehat{\lambda}\left(C-\frac12\right),
\qquad
\widehat{\mu}_2
=
C-\widehat{\lambda}u,
\]
and
\[
\widehat{\mu}
=
\frac{\widehat{\mu}_1+\widehat{\mu}_2}{2},
\qquad
\widehat e
=
\frac{\widehat{\mu}_1-\widehat{\mu}_2}{2}.
\]
\checkmark

\medskip
\noindent Since the six cases are exhaustive, and in each case the parameters
$(\widehat{\lambda},\widehat{\mu},\widehat e)$ satisfy
\eqref{eq:opt_mu}-\eqref{eq:opt_e} and \eqref{eq:opt_lambda}, the
round-clamp abstract transformer is a tight overapproximation. Soundness
then follows immediately from Lemma~\ref{lem:soundness}.
\end{proof}

\section{Soundness and Optimality of the Constrained
Round-Clamp Transformer}
\label{app:constrained_soundness}

The previous section proves that the affine strip
\[
y \in 
[\widehat{\lambda}x+\widehat{\mu}-\widehat e,\,
 \widehat{\lambda}x+\widehat{\mu}+\widehat e]
\]
is sound and optimally tight in width. We now show that the additional
linear constraints used for constrained zonotopes are sound refinements.
Moreover, when the induced polygon has all its vertices on the closed
graph of the round-clamp function, the constrained abstraction is the
tightest possible convex relaxation.

\subsection*{Definitions}

Because $f(x)=\mathrm{clamp}(\lfloor x\rceil,0,C)$ has jumps, we work
with the closed graph
\[
\overline{G}_f([l,u])
=
\overline{\{(x,f(x))\mid x\in[l,u]\}},
\]
where the closure adds one-sided limiting points at rounding thresholds.
At interior rounding thresholds, the closure includes both one-sided limits. At an interval endpoint, it contains the ties-to-even value of $f$ and, when the endpoint is a rounding threshold, also the one-sided limit from inside $[l,u]$.\\

\noindent The affine abstraction defines the strip
\[
P_0
=
\left\{
(x,y)
\;\middle|\;
l\leq x\leq u,\ 
\widehat{\lambda}x+\widehat{\mu}-\widehat e
\leq y
\leq
\widehat{\lambda}x+\widehat{\mu}+\widehat e
\right\}.
\]
The constrained abstraction refines this strip by adding linear cuts
$h_j(x,y)\geq0$:
\[
P_C
=
P_0
\cap
\bigcap_{j=1}^q
\{(x,y)\mid h_j(x,y)\geq0\}.
\]

\noindent Since $P_0$ is the intersection of affine halfspaces, and each
constraint $h_j(x,y)\geq0$ is also an affine halfspace, $P_C$ is a
closed convex polygon.

\begin{definition}[Convex-hull optimality]
The constrained abstraction is convex-hull optimal on $[l,u]$ if
\[
P_C
=
\operatorname{conv}\bigl(\overline{G}_f([l,u])\bigr).
\]
Equivalently, $P_C$ is the smallest convex set containing the closed
graph of $f$ on $[l,u]$.
\end{definition}

\begin{lemma}[Vertex criterion for convex-hull optimality]
\label{lem:vertex_criterion}
Let $P_C\subseteq\mathbb{R}^2$ be a convex polygon. Suppose that
\[
\overline{G}_f([l,u])\subseteq P_C
\]
and
\[
\operatorname{Vert}(P_C)
\subseteq
\overline{G}_f([l,u]).
\]
Then
\[
P_C
=
\operatorname{conv}\bigl(\overline{G}_f([l,u])\bigr).
\]
\end{lemma}

\begin{proof}
Since $P_C$ is convex and contains $\overline{G}_f([l,u])$,
\[
\operatorname{conv}\bigl(\overline{G}_f([l,u])\bigr)
\subseteq
P_C.
\]
Conversely, because $P_C$ is a polygon,
\[
P_C
=
\operatorname{conv}\bigl(\operatorname{Vert}(P_C)\bigr).
\]
By assumption,
\[
\operatorname{Vert}(P_C)
\subseteq
\overline{G}_f([l,u]),
\]
so
\[
P_C
\subseteq
\operatorname{conv}\bigl(\overline{G}_f([l,u])\bigr).
\]
The two inclusions imply equality.
\end{proof}

\noindent\emph{Remark (Degenerate boundaries).}
The oblique slope expressions below are valid
for positive denominators. When $d_l=0$,
we have $l=r_l+\tfrac12$. Under ties-to-even
rounding, both $(l,r_l)$ and $(l,r_l+1)$ belong
to the closed graph, while the corresponding
constraint reduces to $x\geq l$.

Similarly, when $d_u=0$, we have
$u=r_u-\tfrac12$. Both $(u,r_u-1)$ and
$(u,r_u)$ belong to the closed graph, and
the corresponding constraint reduces to $x\leq u$.

Thus, in either case, the degenerate oblique
edge is replaced by an existing vertical
input boundary whose endpoints lie on the
closed graph. The remaining vertices are
unchanged, up to coincidences, so the vertex
criterion continues to hold.

\begin{theorem}[Constrained soundness and optimality] \label{thm:constrained_soundness_optimality} For each round-clamp case, let $P_C$ be the polygon in the $(x,y)$-plane defined by the affine strip together with the constrained-zonotope inequalities listed in Appendix~\ref{app:round_clamp}. Then, 
\[ P_C = \operatorname{conv}\bigl(\overline{G}_f([l,u])\bigr). \] Thus the constrained round-clamp transformer is sound and convex-hull optimal. In particular, no strictly smaller convex relaxation can contain all
concrete graph points in $\overline{G}_f([l,u])$. \end{theorem}

\begin{proof}

The arguments below assume positive denominators.
When a denominator vanishes, the corresponding
constraint reduces to an existing input bound,
as explained in the preceding remark.

We prove the statement by applying Lemma~\ref{lem:vertex_criterion}
case by case. Throughout, $P_C$ is understood to include the input
bounds $l\leq x\leq u$. We also work with the closed graph
$\overline{G}_f([l,u])$, so that both one-sided limiting values at
rounding thresholds are included.

\noindent For each case, it is enough to verify the two conditions of
Lemma~\ref{lem:vertex_criterion}:
\[
\overline{G}_f([l,u])\subseteq P_C,
\]
and
\[
\operatorname{Vert}(P_C)
\subseteq
\overline{G}_f([l,u]).
\]
Both the conditions together imply
\[
P_C
=
\operatorname{conv}\bigl(\overline{G}_f([l,u])\bigr)
\]
by Lemma~\ref{lem:vertex_criterion}.\\

\noindent By Theorem~\ref{thm:main}, the affine strip $P_0$ is sound:
\[
\overline{G}_f([l,u])\subseteq P_0.
\]
Therefore, to prove
\[
\overline{G}_f([l,u])\subseteq P_C
=
P_0
\cap
\bigcap_{j=1}^q
\{(x,y)\mid h_j(x,y)\geq0\},
\]
it is enough to verify that every point of the closed graph satisfies
the additional constrained-zonotope cuts:
\[
h_j(x,y)\geq0
\qquad
\forall (x,y)\in \overline{G}_f([l,u]),
\]
Since $P_C$ is closed, validity on the graph extends to the one-sided
limiting points included in $\overline{G}_f([l,u])$.\\

\noindent Verifying the two conditions proves both convex-hull optimality which encompasses soundness and optimality.

\paragraph{Cases 1-2.}
 In Cases 1 and 2, the function is constant on $[l,u]$. More precisely,
$f(x)=0$ in Case 1 and $f(x)=C$ in Case 2. Hence
\[
\overline{G}_f([l,u])
=
\{(x,C_0)\mid x\in[l,u]\},
\qquad
C_0\in\{0,C\}.
\]
The affine abstraction has $\widehat e=0$ and no additional constrained
cuts are added. Therefore
\[
P_C
=
\{(x,C_0)\mid x\in[l,u]\}
=
\operatorname{conv}\bigl(\overline{G}_f([l,u])\bigr).
\]
Thus the vertex criterion is trivially satisfied. \checkmark

\paragraph{Case 3: $-\tfrac12 \leq l < C+\tfrac12$, $-\tfrac12 < u \leq C+\tfrac12$.}

If $r_l=r_u$, then $f$ is constant on $[l,u]$, and the same argument as
in Cases 1-2 applies:
\[
P_C
=
\{(x,r_l)\mid x\in[l,u]\}
=
\operatorname{conv}\bigl(\overline{G}_f([l,u])\bigr).
\]

\noindent Suppose now that $r_l<r_u$. The affine strip gives the two rounding
envelope inequalities
\[
x-\frac12\leq y\leq x+\frac12.
\]
The four constrained-zonotope inequalities are, in the $(x,y)$-plane,
\[
y\geq r_l,
\qquad
y\leq r_u,
\]
\[
y\leq \widehat{\lambda}_1 x+\widehat{\mu}_1,
\qquad
y\geq \widehat{\lambda}_2 x+\widehat{\mu}_2.
\]
Here
\[
\widehat{\lambda}_1
=
\frac{1}{r_l+\frac12-l},
\qquad
\widehat{\mu}_1
=
r_l-\widehat{\lambda}_1 l,
\]
so the upper oblique cut passes through
\[
(l,r_l)
\quad\text{and}\quad
\left(r_l+\frac12,r_l+1\right).
\]
Similarly,
\[
\widehat{\lambda}_2
=
\frac{1}{u+\frac12-r_u},
\qquad
\widehat{\mu}_2
=
r_u-\widehat{\lambda}_2 u,
\]
so the lower oblique cut passes through
\[
\left(r_u-\frac12,r_u-1\right)
\quad\text{and}\quad
(u,r_u).
\]

\noindent Therefore the boundary of $P_C$ is formed by the following graph
points:
\[
(l,r_l),
\quad
\left(r_l+\frac12,r_l\right),
\quad
\left(r_l+\frac12,r_l+1\right),
\]
\[
\left(r_u-\frac12,r_u-1\right),
\quad
\left(r_u-\frac12,r_u\right),
\quad
(u,r_u),
\]
(vertices listed for the generic configuration; see the remark above). Each of these points belongs to $\overline{G}_f([l,u])$. Moreover, the graph satisfies
all the displayed inequalities. The horizontal cuts hold because \[ r_l\leq f(x)\leq r_u. \] 

\noindent The left oblique cut is valid because
$L_1(x)=\widehat{\lambda}_1x+\widehat{\mu}_1$ passes through
$(l,r_l)$ and $(r_l+\frac12,r_l+1)$, and
$\widehat{\lambda}_1\geq1$. Hence $L_1(x)\geq r_l$ on the first plateau
and $L_1(x)\geq x+\frac12$ after $x=r_l+\frac12$. Since
$f(x)\leq x+\frac12$, we get $f(x)\leq L_1(x)$ on $[l,u]$.

\noindent The right oblique cut is valid because
$L_2(x)=\widehat{\lambda}_2x+\widehat{\mu}_2$ passes through
$(r_u-\frac12,r_u-1)$ and $(u,r_u)$, and
$\widehat{\lambda}_2\geq1$. Hence $L_2(x)\leq x-\frac12$ before
$x=r_u-\frac12$ and $L_2(x)\leq r_u$ on the last plateau. Since
$f(x)\geq x-\frac12$, we get $f(x)\geq L_2(x)$ on $[l,u]$. \checkmark

\paragraph{Case 4: $l \leq -\tfrac12$,
$u \geq C+\tfrac12$, $(l,u)\neq(-\tfrac12,C+\tfrac12)$.}

In this case the closed graph spans the lower clamp, the rounding
staircase, and the upper clamp. Define
\[
\widehat{\lambda}_1
=
\frac{C}{u-\frac12},
\qquad
\widehat{\lambda}_2
=
\frac{C}{C-l-\frac12}.
\]
The lower oblique hull edge is
\[
L(x)=\widehat{\lambda}_1\left(x-\frac12\right),
\]
which passes through
\[
\left(\frac12,0\right)
\quad\text{and}\quad
(u,C).
\]
The upper oblique hull edge is
\[
U(x)=\widehat{\lambda}_2(x-l),
\]
which passes through
\[
(l,0)
\quad\text{and}\quad
\left(C-\frac12,C\right).
\]

\noindent The constrained-zonotope inequalities, together with the affine strip,
induce the polygon
\[
P_C=\{(x,y)\mid l\leq x\leq u,\ 0\leq y\leq C,\ 
L(x)\leq y\leq U(x)\}.
\]

\noindent The four vertices of this polygon are exactly
\[
(l,0),
\qquad
\left(\frac12,0\right),
\qquad
\left(C-\frac12,C\right),
\qquad
(u,C).
\]
Each belongs to $\overline{G}_f([l,u])$.\\

\noindent It remains to check the first condition. By Theorem~\ref{thm:main}, the
closed graph already lies in the affine strip, so it is enough to verify
the additional cuts. The horizontal cuts hold because
\[
0\leq f(x)\leq C.
\]

\noindent The lower oblique cut is valid because
$L(x)=\widehat{\lambda}_1(x-\frac12)$ passes through
$(\frac12,0)$ and $(u,C)$, and
\[
\widehat{\lambda}_1
=
\frac{C}{u-\frac12}
\leq 1.
\]
Hence $L(x)\leq0$ before $x=\frac12$, $L(x)\leq x-\frac12$ on the
rounding region after $x=\frac12$, and $L(x)\leq C$ on the upper clamp
region. Since $f(x)\geq0$, $f(x)\geq x-\frac12$ in the rounding region,
and $f(x)=C$ on the upper clamp region, we get $f(x)\geq L(x), \forall x\in[l,u]$.\\

\noindent The upper oblique cut is valid because
$U(x)=\widehat{\lambda}_2(x-l)$ passes through
$(l,0)$ and $(C-\frac12,C)$, and
\[
\widehat{\lambda}_2
=
\frac{C}{C-l-\frac12}
\leq 1.
\]
Hence $U(x)\geq0$ on the lower clamp region, $U(x)\geq x+\frac12$ on
the rounding region before $x=C-\frac12$, and $U(x)\geq C$ after
$x=C-\frac12$. Since $f(x)=0$ on the lower clamp region,
$f(x)\leq x+\frac12$ in the rounding region, and $f(x)\leq C$ on the upper clamp region, we get $f(x)\leq U(x), \forall x\in[l,u]$. \checkmark

\paragraph{Case 5: $l < -\tfrac12$,
$-\tfrac12 < u < C+\tfrac12$.}
If $r_u=0$, then the function is constant equal to $0$ on $[l,u]$,
and the claim reduces to the constant case. Assume therefore that
$r_u\geq1$.\\

\noindent Define the upper oblique hull edge
\[
U(x)=\widehat{\lambda}x+\widehat{\mu}_1,
\]
where
\[
\widehat{\lambda}
=
\frac{r_u}{r_u-l-\frac12},
\qquad
\widehat{\mu}_1=-\widehat{\lambda}l.
\]
The line $U$ passes through
\[
(l,0)
\quad\text{and}\quad
\left(r_u-\frac12,r_u\right).
\]
Define also the final lower oblique hull edge
\[
L^*(x)=\widehat{\lambda}^*x+\widehat{\mu}^*,
\]
where
\[
\widehat{\lambda}^*
=
\frac{1}{u-r_u+\frac12},
\qquad
\widehat{\mu}^*
=
r_u-\widehat{\lambda}^*u.
\]
The line $L^*$ passes through
\[
\left(r_u-\frac12,r_u-1\right)
\quad\text{and}\quad
(u,r_u).
\]

\noindent The constrained-zonotope inequalities, together with the affine strip, induce the polygon
\[
\begin{aligned}
P_C=\{(x,y)\mid\;& l\leq x\leq u,\ 0\leq y\leq r_u,\ 
x-\tfrac12\leq y,\\
& L^*(x)\leq y\leq U(x)\}.
\end{aligned}
\]

\noindent The vertices of this polygon are exactly
\[
(l,0),
\qquad
\left(\frac12,0\right),
\qquad
\left(r_u-\frac12,r_u-1\right),
\]
\[
(u,r_u),
\qquad
\left(r_u-\frac12,r_u\right),
\]
(vertices listed for the generic configuration; see the remark above). Each belongs to
$\overline{G}_f([l,u])$.\\

\noindent It remains to check the first condition. By Theorem~\ref{thm:main}, the
closed graph already lies in the affine strip, so it is enough to verify
the additional cuts. The horizontal cuts hold because
\[
0\leq f(x)\leq r_u.
\]
The rounding-envelope cut holds because
\[
f(x)\geq x-\frac12.
\]

\noindent The final lower oblique cut is valid because
$L^*(x)=\widehat{\lambda}^*x+\widehat{\mu}^*$ passes through
$(r_u-\frac12,r_u-1)$ and $(u,r_u)$, and
\[
\widehat{\lambda}^*
=
\frac{1}{u-r_u+\frac12}
\geq1.
\]
Hence $L^*(x)\leq x-\frac12$ before $x=r_u-\frac12$ and
$L^*(x)\leq r_u$ on the last plateau. Since
$f(x)\geq x-\frac12$ before the last plateau and $f(x)=r_u$ on the last
plateau, we get $f(x)\geq L^*(x), \forall x\in[l,u]$.\\

\noindent The upper oblique edge is valid because
$U(x)=\widehat{\lambda}x+\widehat{\mu}_1$ passes through
$(l,0)$ and $(r_u-\frac12,r_u)$, and
\[
\widehat{\lambda}
=
\frac{r_u}{r_u-l-\frac12}
\leq1.
\]
Hence $U(x)\geq0$ on the zero plateau, $U(x)\geq x+\frac12$ before
$x=r_u-\frac12$, and $U(x)\geq r_u$ after $x=r_u-\frac12$. Since
$f(x)=0$ on the zero plateau, $f(x)\leq x+\frac12$ before the last
plateau, and $f(x)\leq r_u$ on the last plateau, we get
$f(x)\leq U(x), \forall x\in[l,u]$. \checkmark

\paragraph{Case 6: $-\tfrac12 < l < C+\tfrac12$, $u > C+\tfrac12$.}
If $r_l=C$, then the function is constant equal to $C$ on $[l,u]$,
and the claim reduces to the constant case. Assume therefore that
$r_l<C$.\\

\noindent Define the lower oblique hull edge
\[
L(x)=\widehat{\lambda}x+\widehat{\mu}_2,
\]
where
\[
\widehat{\lambda}
=
\frac{C-r_l}{u-r_l-\frac12},
\qquad
\widehat{\mu}_2=C-\widehat{\lambda}u.
\]
The line $L$ passes through
\[
\left(r_l+\frac12,r_l\right)
\quad\text{and}\quad
(u,C).
\]
Define also the initial upper oblique hull edge
\[
U^*(x)=\widehat{\lambda}^*x+\widehat{\mu}^*,
\]
where
\[
\widehat{\lambda}^*
=
\frac{1}{r_l-l+\frac12},
\qquad
\widehat{\mu}^*
=
r_l-\widehat{\lambda}^*l.
\]
The line $U^*$ passes through
\[
(l,r_l)
\quad\text{and}\quad
\left(r_l+\frac12,r_l+1\right).
\]

\noindent The constrained-zonotope inequalities, together with the affine strip,
induce the polygon
\[
\begin{aligned}
P_C=\{(x,y)\mid\;& l\leq x\leq u,\ r_l\leq y\leq C,\ 
y\leq x+\tfrac12,\\
& L(x)\leq y\leq U^*(x)\}.
\end{aligned}
\]

\noindent The vertices of this polygon are exactly
\[
(l,r_l),
\qquad
\left(r_l+\frac12,r_l\right),
\qquad
(u,C),
\]
\[
\left(C-\frac12,C\right),
\qquad
\left(r_l+\frac12,r_l+1\right),
\]
(vertices listed for the generic configuration; see the remark above). Each belongs to
$\overline{G}_f([l,u])$.\\

\noindent It remains to check the first condition. By Theorem~\ref{thm:main}, the
closed graph already lies in the affine strip, so it is enough to verify
the additional cuts. The horizontal cuts hold because
\[
r_l\leq f(x)\leq C.
\]
The rounding-envelope cut holds because
\[
f(x)\leq x+\frac12.
\]

\noindent The initial upper oblique cut is valid because
$U^*(x)=\widehat{\lambda}^*x+\widehat{\mu}^*$ passes through
$(l,r_l)$ and $(r_l+\frac12,r_l+1)$, and
\[
\widehat{\lambda}^*
=
\frac{1}{r_l-l+\frac12}
\geq1.
\]
Hence $U^*(x)\geq r_l$ on the first plateau and
$U^*(x)\geq x+\frac12$ after $x=r_l+\frac12$. Since
$f(x)\leq x+\frac12$ after the first plateau and $f(x)=r_l$ on the
first plateau, we get $f(x)\leq U^*(x), \forall x\in[l,u]$.\\

\noindent The lower oblique edge is valid because
$L(x)=\widehat{\lambda}x+\widehat{\mu}_2$ passes through
$(r_l+\frac12,r_l)$ and $(u,C)$, and
\[
\widehat{\lambda}
=
\frac{C-r_l}{u-r_l-\frac12}
\leq1.
\]
Hence $L(x)\leq r_l$ before $x=r_l+\frac12$,
$L(x)\leq x-\frac12$ after $x=r_l+\frac12$, and $L(x)\leq C$ on the
upper clamp region. Since $f(x)\geq r_l$ before the first jump,
$f(x)\geq x-\frac12$ in the rounding region, and $f(x)=C$ on the upper
clamp region, we get $f(x)\geq L(x), \forall x\in[l,u]$. \checkmark

\medskip
\noindent Since the six cases are exhaustive, the equality
\[
P_C
=
\operatorname{conv}\bigl(\overline{G}_f([l,u])\bigr)
\]
holds in every case. Therefore the constrained round-clamp transformer
is sound and convex-hull optimal. In particular, no strictly smaller
convex relaxation can contain all concrete graph points.

\end{proof}

\end{document}